\documentclass[10pt]{article}

\usepackage[accepted]{tmlr}

\usepackage{amsmath, amssymb, amsfonts}
\usepackage{booktabs}
\usepackage{graphicx}
\usepackage{caption}
\usepackage{subcaption}
\usepackage{multirow}
\usepackage{overpic}
\usepackage{xcolor}
\usepackage{xspace}
\usepackage{pifont}
\usepackage{url}
\usepackage{hyperref}
\usepackage[nameinlink,noabbrev]{cleveref}
\hypersetup{hidelinks}

\newcommand{\cmark}{\textcolor{teal}{\ding{51}}}
\newcommand{\xmark}{\textcolor{red}{\ding{55}}}
\newcommand{\model}{PCHMR\xspace}

\title{PCHMR: Empowering and Benchmarking Human Mesh Recovery in Privacy-Constrained Real-World Settings}

\author{\name Zeyu Cao\thanks{Equal contribution.} \email zeyu.cao@cl.cam.ac.uk \\
\addr University of Cambridge
\AND
\name Qingxuan Wu\footnotemark[1] \email qingxuan.wu1@gmail.com \\
\addr University of Pennsylvania
\AND
\name Zhiyang Dou\thanks{Co-corresponding authors.} \email zhiyang0@connect.hku.hk \\
\addr The University of Hong Kong
\AND
\name Rui Xu \email ruixu1999@connect.hku.hk \\
\addr The University of Hong Kong
\AND
\name Yuan Liu \email yuanly@ust.hk \\
\addr Hong Kong University of Science and Technology
\AND
\name Javier Fernandez-Marques \email javier@flower.ai \\
\addr Flower Labs
\AND
\name Nicholas D. Lane \email ndl32@cam.ac.uk \\
\addr University of Cambridge
\AND
\name Taku Komura \email taku@cs.hku.hk \\
\addr The University of Hong Kong
\AND
\name Wenping Wang\footnotemark[2] \email wenping@tamu.edu \\
\addr Texas A\&M University}

\def\month{10}
\def\year{2026}
\def\openreview{\url{https://openreview.net/forum?id=4XHzBaL0qA}}

\begin{document}
\maketitle

\begin{abstract}
Fine-tuning directly on user-side data is an effective way to improve the performance of widely adopted data-driven human mesh recovery (HMR) models. However, doing so for existing HMR methods often assumes aggregating user-side data from real-world HMR-deployed devices to a central training server, posing a significant privacy risk as sensitive human images are transmitted. How can human mesh recovery be evaluated and improved in privacy-constrained real-world settings? We study this question through a benchmark in which state-of-the-art HMR models are trained with federated learning and a secure-aggregation variant, neither of which centralizes raw images, and evaluated across a wide range of client-scale and data-heterogeneity settings. We document that common HMR training pipelines are built around centralized data access and quantify how representative HMR backbones behave when that assumption is removed. Furthermore, to study the data bottleneck of privacy-constrained HMR training, we propose DePoser, a local annotation and fine-tuning pipeline enhanced with depth foundation models, with which collaboratively trained HMR models can be locally tailored to the end user's distribution. On in-the-wild data, fine-tuning on DePoser annotations improves personalization accuracy, measured against DePoser-generated pseudo-ground truth. This benchmark aims to support future work on privacy-constrained HMR models and their real-world deployment and evaluation.

\end{abstract}
\vspace{-4mm}

\section{Introduction}
\begin{figure}[t]
\vspace{-4mm}
 \centering
 \includegraphics[width=0.7\linewidth]{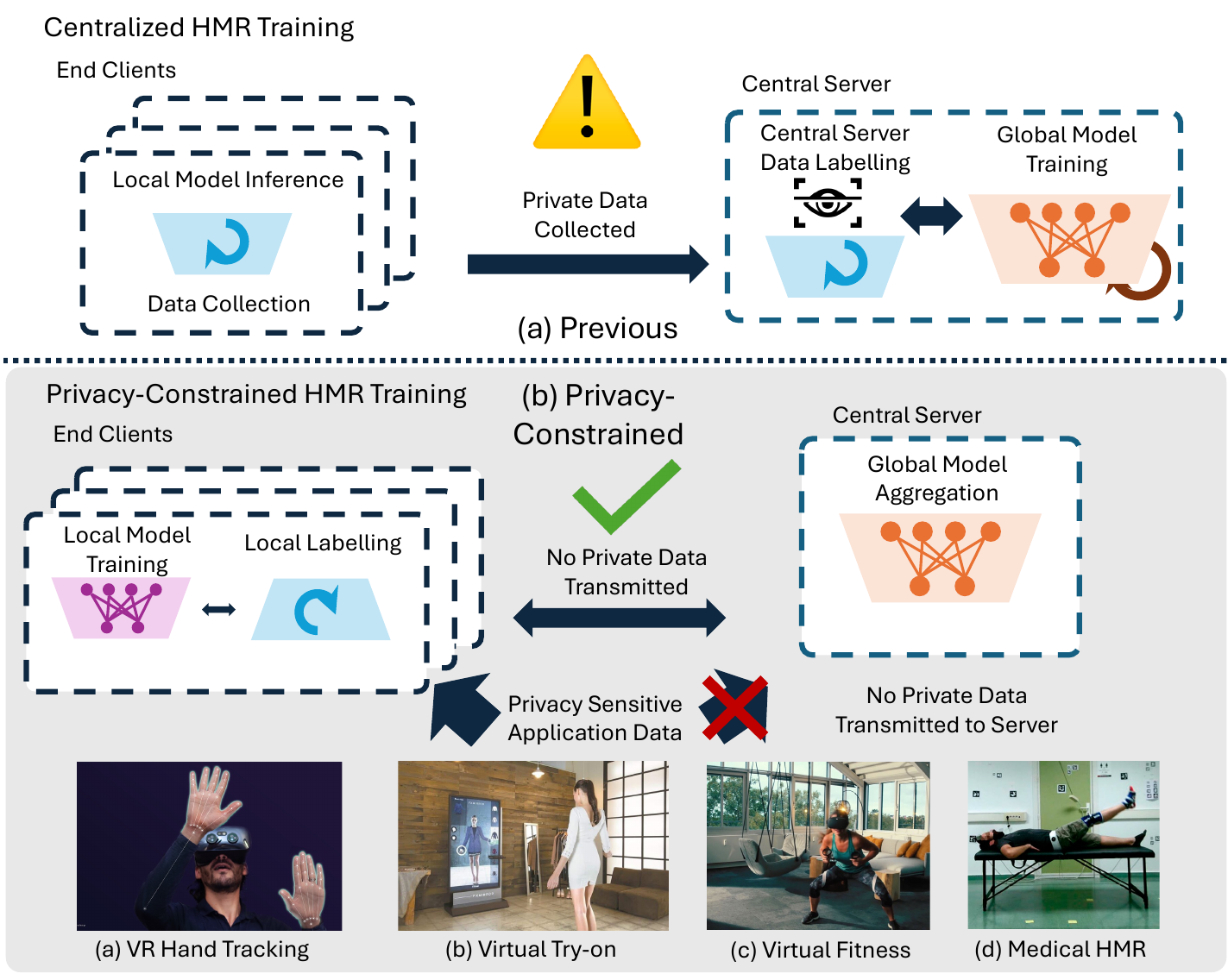}
 \captionsetup{font=small,labelfont=bf}
 \caption{\textbf{Comparison between centralized and privacy-constrained HMR frameworks.} Existing HMR frameworks commonly assume centralized training (a), which requires uploading sensitive user data and therefore raises privacy concerns. In comparison, \model benchmarks privacy-constrained settings for HMR models (b), where raw data remain on the device and only model updates are exchanged. Locally generated annotations then provide additional supervision, so HMR models can be improved with \textit{privacy-constrained real-world human data}.
 }
 \label{fig:pf-hmr-overview}
\end{figure}

Advancements in human mesh recovery (HMR) methods~\citep{lin2021end, cho2022cross, dou2023tore, li2023niki, lin2021mesh, li2021hybrik} are relevant to a wide range of real-world applications, such as virtual reality~\citep{apple2024apple, meta2024meta, vrchat2024vrchat}, augmented reality~\citep{meta2024project}, human behavior analysis~\citep{liu2022close, guo2023student, zhang2023popularization, zhang2024analysis}, and virtual try-on~\citep{fxgear2024vr}. Accurate human mesh recovery is always desired for those human-centric applications.

For general learning-based applications like medical image computing~\citep{guan2024federated, adnan2022federated, rehman2023federated} or on-device speech recognition~\citep{brydinskyi2024enhancing, azam2023importance, azam2023federated},
fine-tuning the model \textit{directly on user-side data} is one of the most effective ways of enhancing the model performance by minimizing the domain gap~\citep{mairittha2020improving, lee2023fine}. To the best of our knowledge, existing HMR frameworks~\citep{kanazawa2018end, lin2021end, dou2023tore, cho2022cross, choi2020pose2mesh, li2021hybrik, li2025hybrik, li2023niki, lin2023one} are typically developed and evaluated under the setting where all training data are on a single device, which we refer to as the \textit{centralized} setting. As a result, HMR model training using existing frameworks usually assumes the collection and aggregation of sensitive data from user devices to a central training server. This raises privacy and compliance concerns and is increasingly challenging, if not infeasible, under strict privacy regulations such as the General Data Protection Regulation (GDPR)~\citep{european2016regulation}.
A na\"ive raw-data-local solution would be to train the HMR model on each edge device using only the scarce data available there. However, since the mapping from a monocular image to a 3D human's pose and shape is highly complex, training such a model with limited real-world data collected from only a single edge device would result in significantly inferior performance compared to centralized training; see Appendix~\ref{app:naive-single-device-training}. This poses a \textit{performance-privacy dilemma}: the desire to effectively utilize user data on the edge to enhance the model's performance, conflicting with the principle of avoiding raw-data centralization.

Despite extensive work on HMR~\citep{kanazawa2018end, lin2021end, dou2023tore, cho2022cross, choi2020pose2mesh, li2021hybrik, li2025hybrik, li2023niki, lin2023one, fiche2025mega} where models are trained and evaluated centrally under ideal conditions, one practical question remains unexplored: \textit{How do state-of-the-art methods perform under privacy-constrained real-world settings, where centralized aggregation of data from user devices is infeasible due to privacy concerns?}

This work presents a systematic benchmark study of HMR models~\citep{dou2023tore, cho2022cross}\footnote{Although we select two representative transformer-based methods, the framework discussed in this paper is agnostic to the choice of HMR architecture.} under a comprehensive series of privacy-constrained, real-world settings, answering the aforementioned question. In all of our experiments, the training is conducted with a client-server model, where each client holds a part of the training data, which is never shared. Under this framework, we then scale up to a large number of clients, where we find that the HMR models converge with minimal accuracy loss even with 1,000 clients. In addition, since each HMR-equipped device user could have a different usage pattern, we simulate real-world data non-uniformity with three schemes of varying complexity. When data heterogeneity is pronounced, collaboratively trained HMR models can match, and in some settings slightly exceed, their centrally trained counterparts. We interpret this as evidence that federated training need not degrade HMR accuracy under the tested protocols; we discuss possible causes of the occasional gains in Appendix~\ref{app:training-budget}.
Furthermore, we evaluate a secure-aggregation variant with MPC~\citep{knott2021crypten}. This variant protects the aggregation channel under an honest-but-curious server assumption and prevents the server from directly observing individual client updates, but it should not be read as an attack-based privacy evaluation of the released global HMR model. With less than $2\times$ the runtime in our small-scale measurement, MPC appears feasible enough to merit further system-level study.

In privacy-constrained HMR, raw images cannot be transmitted to a central server, so locally annotated data becomes essential both for collaborative training and for model personalization. Because user- and environment-specific scenarios vary widely, the annotation pipeline must be robust; optimization-based annotators suit this need because they are stable and accurate. Unlike existing annotators~\citep{bogo2016keep, pavlakos2019expressive} that rely on sparse cues such as 2D keypoints or masks, whose performance can be limited by detector quality and monocular ambiguity, we introduce \textit{DePoser}, a depth-aware HMR annotation pipeline.

DePoser leverages the large human vision foundation model Sapiens~\citep{khirodkar2024sapiens}, a 2B-parameter model trained on 300M+ in-the-wild images, to provide body-part relative depth cues, and then aligns these predictions to metric scale via a lightweight linear regressor trained with a general absolute depth estimator, preserving Sapiens' structure. At the same time, we exploit Sapiens' accurate and robust 2D joint predictions to provide stronger optimization signals. This yields 2.5D keypoints that couple precise localization with scale-invariant depth. DePoser is fully monocular, deployable on common devices from a single RGB image input. We validate our pipeline against~\citet{pavlakos2019expressive} and benchmark downstream HMR models on a VR dataset of internet-sourced and in-house footage annotated by DePoser.

We show that locally fine-tuning models on such DePoser-generated annotations enhances HMR performance in privacy-constrained collaborative training setups, enabling on-device models to self-collect pseudo-supervision. In summary, our contributions are:
\begin{itemize}
 \item We identify the centralized-data assumption in state-of-the-art HMR training pipelines and evaluate how these models perform without it, which is essential for deploying them on private real-world data.
 \item We provide a comprehensive study of state-of-the-art HMR models, investigating their performance under privacy-constrained training settings of varying complexity and assessing the feasibility of collaborative training for HMR models.
 \item Local annotators are essential for privacy-constrained HMR training. Motivated by the depth ambiguity issue of previous optimization-based HMR annotators, we further propose DePoser, a local HMR annotation pipeline enhanced by depth foundation models, which can be integrated with the proposed collaborative training framework while making the pseudo-label dependency explicit.
\end{itemize}

\section{Related Work}
Many recent works study 3D representations of the human body or body parts, such as 3D skeleton~\citep{martinez2017simple, mehta2017vnect, yang20183d}, 3D whole-body mesh~\citep{li2021hybrik, li2023niki, dou2023tore, cho2022cross, choi2020pose2mesh, lin2021end}, 3D hand/face recovery~\citep{wu2025dice, shimada2023decaf, moon2024dataset}, and 3D human motion generation~\citep{guo2022generating, tevet2022human, chen2023executing, guo2024momask, liang2024intergen, wu2026text2interact, javed2024intermask, zhao2026comovi}. Due to the large amount of prior work, we focus on 3D whole-body mesh recovery in the following section for its high relevance.
\subsection{Human Mesh Recovery}
\textbf{Optimization-based Human Mesh Recovery.}
This approach estimates the 3D mesh that is consistent with the 2D observation under various constraints by directly optimizing with standard body model parameters. After the seminal work SMPLify~\citep{bogo2016keep}, follow-up work extends the fitting objective with additional terms that keep the 3D mesh consistent with the 2D image observation. In addition, factors such as shape, pose, and segmentation provide additional priors to the optimization problem, such as HuMoR~\citep{rempe2021humor}, Pose-NDF~\citep{tiwari2022pose} and others~\citep{guler2018densepose, ke2022modnet}. This type of method has also been expanded to regress detailed human body, hand, and face at the same time~\citep{pavlakos2019expressive, zioulis2023kbody, lin2024motion}. Optimization-based methods apply gradient-based optimization with a neural network and perform iteratively to obtain accurate mesh results such as HUND~\citep{zanfir2021neural} and LVD~\citep{corona2022learned}. These methods generally yield a higher quality of mesh compared with the regression-based approach; however, due to their iterative optimizing nature, these methods are often incapable of real-time inference.

\noindent
\textbf{Regression-based Human Mesh Recovery.}
This method generates a 3D human mesh directly from pixels with an end-to-end network.
Various methods use intermediate representations such as joints~\citep{xiang2019monocular}, segmentation~\citep{rueegg2020chained}, UV maps~\citep{yao2019densebody}, heatmaps~\citep{moon2020i2l}, keypoints and silhouettes~\citep{pavlakos2018learning}, dense correspondences~\citep{xu2019denserac}, or texture consistency~\citep{pavlakos2019texturepose} to supply additional information for the mesh regression. In addition, forward kinematics and inverse kinematics may be considered at the same time as additional constraints for the neural network, such as HybrIK~\citep{li2021hybrik}, HybrIK-X~\citep{li2025hybrik}, and NIKI~\citep{li2023niki}. This representation is then passed into a decoder network, the output of which could be classified into the following two categories: \textit{parametric output} and \textit{non-parametric output}. The former is in the parameter space of parametric models such as SMPL~\citep{loper2015smpl}, while the latter is in 3D space, such as joints and mesh vertex coordinates. SOTA performance has been achieved by transformer-based methods such as METRO~\citep{lin2021end}, MeshGraphormer~\citep{lin2021mesh}, and FastMETRO~\citep{cho2022cross}, which are non-parametric approaches excelling at modeling long-range dependency, making them robust even under occlusions. Transformer efficiency is optimized with token pruning methods such as in TORE~\citep{dou2023tore}.

\subsection{Federated Learning}

Federated Learning~(FL)~\citep{mcmahan2017communication} is a family of distributed learning paradigms in which raw training data never leave the clients; we refer to the conventional alternative as \textit{centralized training}. Each client instead trains a local model and sends its updates to a central server, which \textit{aggregates} them into a global model. FL reduces raw-data exposure but does not by itself guarantee privacy: model updates and the released global model can still leak information~\citep{kairouz2021advances}, which is why FL is often combined with secure aggregation or differential privacy.

FL has been applied to recommender systems~\citep{du2021federated, duan2020jointrec, muhammad2020fedfast}, finance~\citep{cheng2021secureboost}, edge computing~\citep{ye2020edgefed, jiang2020customized, wang2021efficient, liu2021blockchain}, industrial applications~\citep{tang2022federated, zhao2020intelligent, li2020deepfed}, and healthcare~\citep{ouyang2021novel, dayan2021federated, ma2022assisted, chen2020fl} thanks to its collaborative and raw-data-local nature. FL also demonstrates strong performance in computer vision~\citep{liu2020fedvision, li2019privacy}, natural language processing~\citep{yang2018applied, lin2022fednlp}, and automatic speech recognition~\citep{azam2023federated, zhu2022decoupled}. However, to our knowledge, it has not yet been used to address data grounding challenges in real-world monocular HMR. Existing FL-based computer vision solutions cover tasks such as object detection~\citep{liu2020fedvision} and segmentation~\citep{li2019privacy}, and human-related FL work has addressed pose estimation and person re-identification~\citep{zhuang2024coala} as well as posture recognition~\citep{wang2024multi}, tasks that differ significantly from monocular 3D HMR. In contrast, our approach directly regresses the 3D human mesh from standard RGB images, offering a more versatile and practical solution without the need for specialized hardware.

Other distributed training paradigms are also relevant. Split learning trains different model segments across client and server and exchanges intermediate activations rather than complete model updates~\citep{vepakomma2018split}, while FedSplit studies operator-splitting-based federated optimization~\citep{pathak2020fedsplit}. These approaches may be useful for future HMR systems with different communication, compute, and leakage trade-offs. In this paper, we focus on federated parameter sharing because it directly tests whether existing HMR backbones can be trained without centralizing raw user images, and because it is readily supported by mature benchmark infrastructure.

\section{Method}

\noindent
Central to the \model framework is an HMR model $\mathcal{M}$ with parameters $w$, which takes a monocular image $\mathcal{I} \in \mathbb{R}^{H \times W \times 3}$ as input and outputs a 3D human mesh $\mathcal{V} \in \mathbb{R}^{6890 \times 3}$ in the SMPL~\citep{loper2015smpl} topology.

\begin{figure}[t]
\centering
\vspace{-7mm}
\begin{overpic}
[width=\linewidth]{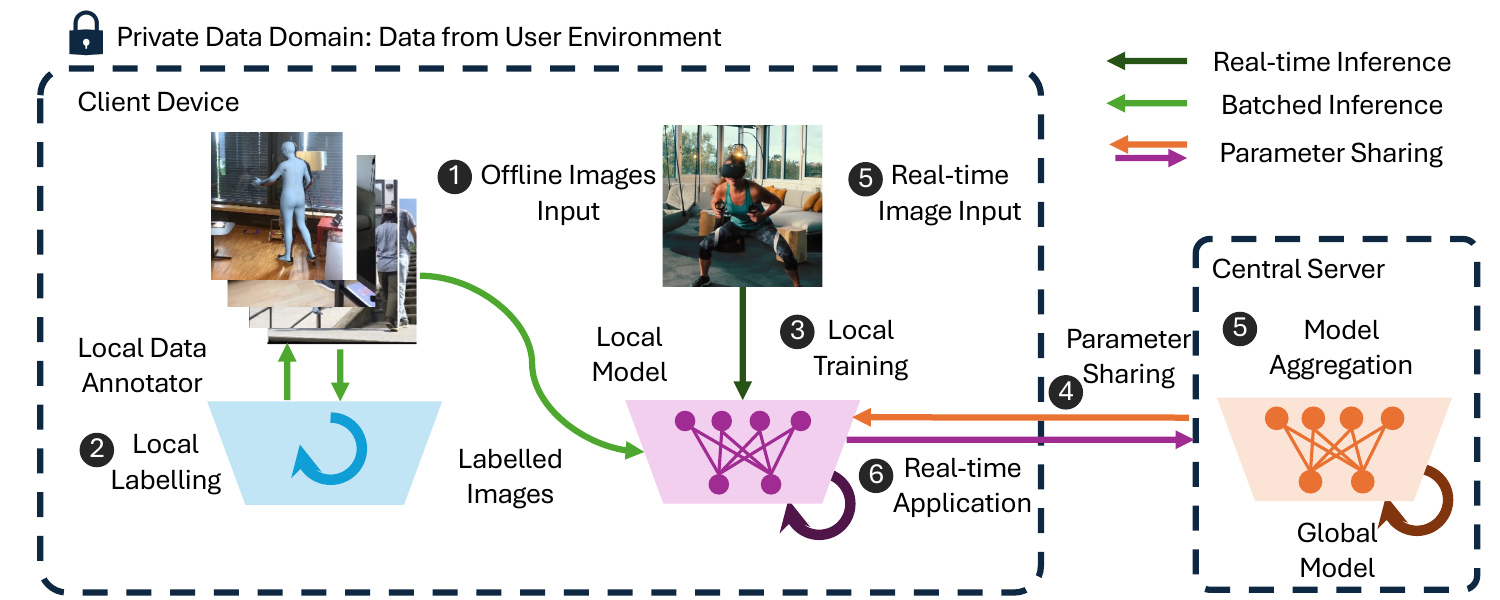}
\end{overpic}
\captionsetup{font=small,labelfont=bf}
\caption{\textbf{Overview of the Pipeline for Privacy-Constrained HMR.} Data labeling can be done offline via a local data annotator (1, 2). After that, the labels of user-generated data are used to collaboratively train a new global model with privacy constraints (3, 4). User data never leave the device. The global model is then deployed for real-time on-device inference (5, 6). \model benchmarks implementations of such a privacy-constrained pipeline under various real-world settings. }
\label{fig:pipeline}
\vspace{-3mm}
\end{figure}

Under our privacy-constrained setting, the HMR model is trained on a distributed HMR dataset, divided into disjoint subsets $\mathcal{D} = \cup_{k=1}^{n} \mathcal{D}_k$ and $\mathcal{D}_i \cap \mathcal{D}_j = \emptyset$ $\forall i \neq j$, where each subset $\mathcal{D}_k$ is locally collected and annotated on device $k$. Each local dataset consists of $n_k$ locally annotated image-human mesh pairs, $\mathcal{D}_k = \{(\mathcal{I}_i, \mathcal{V}_i)\}_{i=1}^{n_k}$. An overview of the \model framework is shown in \cref{fig:pipeline}.

We first conduct a comprehensive benchmark of HMR in privacy-constrained real-world settings in \cref{sec:benchmark}. Then, in \cref{sec:depth_annotator}, we introduce and evaluate a local data annotator, named DePoser, which enables human mesh annotation for privacy-constrained HMR training.

\subsection{Privacy-Constrained Benchmark Construction}
\label{sec:benchmark}

We create a client-server framework to train HMR models in a raw-data-local way to construct our benchmark. Specifically, we use the Flower~\citep{beutel2020flower} library to simulate a federated training setup. While Secure Multi-Party Computation~(MPC)~\citep{knott2021crypten} and Differential Privacy~(DP)~\citep{dwork2006differential,abadi2016deep} are also considered for privacy-preserving machine learning methods, due to the computation and communication overhead of completely setting up MPC training and the performance degradation of DP under large-scale models (such as SOTA HMR models), we primarily focus on benchmarking under the Federated Learning~\citep{mcmahan2017communication} approach. We consider the basic setup for federated training and the secure-aggregation variant enabled by MPC. Additional details on MPC and DP are discussed in Appendix~\ref{app:mpc_dp}.

\noindent \textbf{Privacy Scope and Threat Model.} Our benchmark evaluates utility when raw images and mesh annotations remain on client devices. In vanilla FL, the server still receives individual model updates, so the protocol reduces raw-data centralization but does not by itself rule out update-level or released-model leakage. In the secure-aggregation variant, we assume an honest-but-curious server and use MPC to prevent the server from observing individual client updates before aggregation. We do not claim formal differential privacy and do not perform membership-inference, gradient-inversion, or model-inversion attacks in this work; these attack-based privacy evaluations are important future extensions.

We conduct a series of comprehensive experiments on two popular HMR models~\citep{cho2022cross,dou2023tore} in various settings that closely reflect the complexities of the real-world training environment:

\noindent \textbf{Large Number of Participating Devices.} In a typical real-world privacy-respecting setup, a large number of user devices are often involved~\citep{huba2022papaya, bonawitz2019towards}. In addition, due to varied usage patterns, it is impossible for each user to participate in every global fitting round. Based on these observations, we perform experiments with 10, 100, and 1000 simulated user devices (clients), with data randomly partitioned for each device to simulate locally collected and stored private training data. The number of our clients is limited by the total training dataset size ($\sim$500,000 samples), giving roughly 500 (1,000 clients) to 50,000 (10 clients) samples per client. In each global training round, 10 clients are randomly selected as participants, each of which trains for one local epoch ($E = 1$), a single pass over its own shard, using the locally partitioned data. Then, the model weights are uploaded to the central server and aggregated, while the local dataset remains local for raw-data locality.

\noindent \textbf{Heterogeneous Data Distribution.} In the privacy-constrained HMR training setup, the training data are collected and annotated locally. Thus, each user's locally annotated dataset can differ substantially in data distribution and volume from another user's local dataset due to variation in human identity, body shape, attire, and usage environment. We employ multiple methods to simulate this highly non-uniform data distribution:

1) We use the Dirichlet-based label partition commonly called Latent Dirichlet Allocation (LDA) in FL~\citep{hsu2019measuring}, with the parameter $\alpha$ to control the degree of heterogeneity of the distribution. See Appendix~\ref{app:lda-data-partition} for a visualization.

2) We create a mixed dataset comprising a series of datasets~\citep{ionescu2013human3, lassner2017unite, lin2014microsoft, andriluka20142d, mehta2018single} with 3DPW~\citep{von2018recovering} and assign clients according to datasets. This creates a more heterogeneous setup, benchmarking the SOTA HMR model's capability under challenging training settings.

3) To simulate real-world fine-tuning, where a centrally pre-trained HMR model is fine-tuned with user-side data, we perform privacy-constrained fine-tuning on the in-the-wild 3DPW~\citep{von2018recovering} dataset. We further adopt the challenging setup of assigning each scene to a participating client, which resembles real-world use cases where each client uses their edge device in a different environment.

\noindent \textbf{Privacy-Constrained Personalized Fine-Tuning.}
Data distribution alignment is essential in real-world deployment of HMR models, which could be effectively achieved by directly fine-tuning on the end user's data distribution.
To evaluate the effectiveness of model personalized fine-tuning, where the federated-trained HMR model is further fine-tuned on each user's local dataset, we collect a dataset of over 3000 images from the internet and in-house VR usage footage, comprising 4 different scenes with users using VR and AR devices. The internet subset is drawn from publicly accessible videos, and the in-house subset is recorded with participant consent for research use. We further randomly partition the dataset into training and validation sets, fine-tuning the model for 2 epochs over the user's local training split and then evaluating on the validation sets.

\subsection{DePoser: Depth-Aware Local Annotator}
\label{sec:depth_annotator}

Personalizing an HMR model requires annotated user-side data, but privacy constraints and the sensitivity of human images rule out collecting such images centrally. Privacy-constrained HMR training therefore needs an accurate annotation pipeline that runs on the device.

Optimization-based HMR methods~\citep{bogo2016keep, pavlakos2019expressive} are suitable candidates for local annotation due to their robustness and accuracy. These methods often use 2D keypoints to guide the optimization process of the human mesh. However, relying solely on 2D keypoints introduces monocular ambiguity, often leading to solutions that produce plausible 2D re-projections but suffer from poor 3D accuracy. Inspired by SMPLify-X~\citep{pavlakos2019expressive} and DICE~\citep{wu2025dice}, we devise a depth-aware monocular local annotation pipeline named DePoser, where the mesh is supervised not only by 2D keypoints but also by scale-invariant depth information in the z-direction.
First, we extract a depth map and 2D keypoints from a large human foundation model, Sapiens~\citep{khirodkar2024sapiens}. By taking advantage of a model
pre-trained on vast in-the-wild human data, our model benefits from a strong human prior, and the fitting objective is driven by 2.5D rather than 2D cues for in-the-wild user-side scenarios.
Second, our method does not require expensive hardware such as RGB-D sensors, LiDAR sensors, or synchronized multi-view cameras, making it possible to deploy DePoser to ordinary consumer-grade devices with only monocular cameras.

Specifically, given an RGB image, we first use Sapiens-pose~\citep{khirodkar2024sapiens} to extract 2D human keypoints. Then, the corresponding pseudo-depth of each keypoint is extracted from a depth map obtained using Sapiens-depth~\citep{khirodkar2024sapiens}. Subsequently, we minimize the following objective function with the L-BFGS optimizer~\citep{nocedal2006numerical}:
\begin{equation}
 E(\beta, \theta)= E_J+\lambda_{\theta_b}E_{\theta_b}
 + \lambda_\alpha E_\alpha+\lambda_\beta E_\beta+\lambda_{\mathcal{C}} E_{\mathcal{C}} + \lambda_{d} E_d,
\end{equation}
where $E_J$ is the 2D joint projection loss, $E_{\theta_b}$ is the pose prior~\citep{pavlakos2019expressive} loss, $E_\alpha=\sum_{i \in(\text {elbows}, \text {knees})} \exp \left(\theta_i\right)$ is the elbow/knee bending regularizer, $E_\beta = \|\beta\|^2$ is the shape regularizer, $E_{\mathcal{C}}$ the interpenetration penalty, and $E_d$ the scale-invariant logarithmic (SILog)~\citep{eigen2014depth} loss:
\begin{equation}
 \label{eq:silog}
 E_d = \sqrt{\;\frac{1}{N} \sum_{i \in \mathcal{J}} d_i^2
 \;-\; \lambda_{\mathrm{si}}\left(\frac{1}{N} \sum_{i \in \mathcal{J}} d_i\right)^{2}\;},
\end{equation}
where $d_i = \log z_i-\log \hat{z}_i$ and $\hat{z}_i, z_i$ are the predicted and pseudo-GT depth values of the $i$-th joint, $\mathcal{J}$ is the set of joints, and $N = |\mathcal{J}|$. Rescaling all predicted depths shifts every $d_i$ by a common constant. Since $E_d^2 = \mathrm{Var}(d) + (1-\lambda_{\mathrm{si}})\,\bar{d}^{\,2}$, the variance term is unaffected and only the down-weighted mean term changes, so $E_d$ reduces sensitivity to the global depth scale and emphasizes the relative depth structure across joints; it is strictly scale-invariant only for $\lambda_{\mathrm{si}}=1$. We use the standard setting $\lambda_{\mathrm{si}} = 0.5$ of~\citet{eigen2014depth}.

\section{Evaluation}
\label{sec:evaluation}

\subsection{Experimental Setup}
\label{sec:experiment-setup}
\noindent
\textbf{Datasets.} Following previous works~\citep{cho2022cross, dou2023tore, lin2021end}, we use the Human3.6M~\citep{ionescu2013human3}, 3DPW~\citep{von2018recovering}, UP-3D~\citep{lassner2017unite}, COCO~\citep{lin2014microsoft}, MPII~\citep{andriluka20142d}, and MuCo-3DHP~\citep{mehta2018single} datasets to train our models. We evaluate our model using the Human3.6M~\citep{ionescu2013human3} and 3DPW~\citep{von2018recovering} datasets.

In addition, we also curate a dataset from internet video footage and in-house VR usage. This dataset consists of over 3000 image-mesh pairs with 4 different scenes and identities. We conduct local fine-tuning and testing experiments on this dataset. For the internet-sourced subset, we use publicly accessible videos and retain source metadata for reproducibility. For the in-house subset, participants consented to research use of the recordings. Since the mesh labels are generated by DePoser rather than captured from an independent multi-view or marker-based system, we report these experiments as pseudo-labeled personalization studies and do not claim independent ground-truth coverage or demographic representativeness. Annotation quality is controlled through the quantitative DePoser comparison on 3DPW in Sec.~\ref{sec:local-annotation-eval} and by excluding cases where the local fitting pipeline fails to produce a valid body fit.

\noindent
\textbf{Evaluation Metrics.}
Following previous works~\citep{cho2022cross, dou2023tore, lin2021end}, we use \textit{Mean Per Joint Position Error} (MPJPE)~\citep{ionescu2013human3, zhou2018monocap} and
\textit{Procrustes-Aligned Mean Per Joint Position Error} (PA-MPJPE)~\citep{pavlakos2018learning} to evaluate our models' accuracy. \textit{Mean Per Vertex Position Error} (MPVPE)~\citep{pavlakos2018learning} is also calculated when a ground-truth 3D mesh is available. All the error metrics are in millimeters (mm).

\begin{sloppypar}
\noindent
\textbf{Simulation.} Following previous works that adopted the federated learning paradigm \citep{jithish2023distributed, lubana2022orchestra, madill2022scalesfl, gao2022end, jiang2023fdapt}, we conduct our experiments in a simulated environment using Flower~\citep{beutel2020flower} to avoid dealing with low-level implementations such as communication and resource scheduling.
\end{sloppypar}

\noindent \textbf{Reproducibility and Release Plan.} All reported client partitions are generated with fixed random seeds and fixed partition metadata. Upon publication we will release code, but no data. The release covers both contributions: the Flower-based \model harness and the DePoser implementation, including their training, evaluation, and deterministic data-splitting code together with the hyperparameter configurations of every reported run. Because the datasets we train and evaluate on~\citep{ionescu2013human3, lassner2017unite, lin2014microsoft, andriluka20142d, mehta2018single, von2018recovering} are the property of their respective rights holders, we redistribute none of them; users obtain them directly from their owners under the original terms, and we supply a guide for preparing them and replicating our results with our code. The VR footage we curated for the personalization study is not released in any form (see the Broader Impact Statement).

\subsection{Privacy-Constrained Evaluations}

We choose two representative SOTA HMR models~\citep{cho2022cross, dou2023tore} for their robustness and accuracy. Note that, due to the well-modularized nature of the HMR community, other models can be easily incorporated into our benchmark pipeline. For the chosen models, we perform extensive experiments in a series of simulated settings that mimic scenarios encountered in real-world privacy-constrained training. We group the experiments into categories and outline each category in the following sections.

\begin{table}[htbp]
\centering
\small
\captionsetup{labelfont=bf}

\resizebox{0.8\columnwidth}{!}{%
\begin{tabular}{lccc|c}
\toprule
\textbf{\#Clients (Total / Per Round}) & \textbf{Model} & \textbf{MPJPE} & \textbf{PA-MPJPE} & \textbf{Raw Data Local?} \\
\midrule
Centralized & FM-S & 57.98 & 40.62 & \xmark\\ \midrule
10 / 10 & FM-S & 56.85 & 40.52 &\cmark\\
100 / 10 & FM-S & 59.31 & 41.48 &\cmark\\
1000 / 10 & FM-S & 60.82 & 44.53 &\cmark \\
\midrule
Centralized & TORE-S & 63.88 & 41.99 & \xmark\\ \midrule
10 / 10 & TORE-S & 61.27 & 41.60 &\cmark\\
100 / 10 & TORE-S & 62.92 & 43.11 &\cmark\\
1000 / 10 & TORE-S & 66.04 & 44.04 &\cmark\\
\bottomrule
\end{tabular}
}

\caption{\textbf{Effect of Number of Clients.} HMR results (MPJPE and PA-MPJPE) for the \model baseline experiments (\texttt{10/10}), the Independent and Identically Distributed (IID) data 100-client experiments (\texttt{100/10}), and IID data 1000-client experiments (\texttt{1000/10}) with their best evaluation results. All experiments select 10 clients per round uniformly at random. For comparison, we also add the centralized baseline trained with the same model and the same dataset. }
\label{tab:basic-fl-eval}
\vspace{-3mm}
\end{table}

\begin{table}[htbp]
\centering
\small
\captionsetup{labelfont=bf}
\resizebox{0.55\linewidth}{!}{%
\begin{tabular}{l|cc|cc}
\toprule
\multicolumn{1}{c|}{\textbf{Optimizer}} & \multicolumn{2}{c|}{\textbf{FM-S Model}} & \multicolumn{2}{c}{\textbf{TORE-S Model}} \\
\multicolumn{1}{c|}{} & \textbf{MPJPE} & \textbf{PA-MPJPE} & \textbf{MPJPE} & \textbf{PA-MPJPE} \\
\midrule
\multicolumn{5}{c}{\textbf{IID Distribution}} \\
\midrule
FedAvg & 59.31 & 41.48 & 62.92 & 43.11 \\
FedOpt & 58.49 & 41.43 & 62.82 & 42.34 \\
FedProx & 57.71 & 41.26 & 63.41 & 43.17 \\
\midrule
\multicolumn{5}{c}{\textbf{LDA $\alpha=0.5$ Distribution}} \\
\midrule
FedAvg & 57.80 & 39.82 & 58.91 & 40.94 \\
FedOpt & 55.41 & 39.36 & 59.34 & 41.92 \\
FedProx & 56.46 & 39.58 & 61.57 & 41.10 \\
\midrule
\multicolumn{5}{c}{\textbf{LDA $\alpha=1$ Distribution}} \\
\midrule
FedAvg & 59.63 & 39.80 & 62.39 & 41.76 \\
FedOpt & 58.10 & 40.10 & 59.57 & 42.10 \\
FedProx & 56.45 & 39.28 & 58.96 & 41.97 \\
\midrule
\multicolumn{5}{c}{\textbf{LDA $\alpha=10$ Distribution}} \\
\midrule
FedAvg & 56.09 & 39.84 & 60.96 & 42.25 \\
FedOpt & 58.25 & 40.47 & 58.83 & 41.29 \\
FedProx & 57.33 & 40.27 & 58.96 & 41.57 \\
\bottomrule
\end{tabular}
}
\caption{\textbf{Effect of Federated Optimizer.} Performance of HMR models with different federated optimizers (FedAvg, FedOpt, FedProx) on IID data and LDA skewed data. The results are measured with 100-client experiments (\texttt{100/10}) with 10 clients sampled per round.}
\label{tab:fed-optimizers}
\end{table}

\subsubsection{Performance on a Large Number of Clients}

We evaluate the performance of HMR models when trained under a simulated Federated Learning environment with 10, 100, and 1000 participating devices, where 10 devices are randomly selected to participate at each global round. The environment has a client-server configuration, where each client trains the model for one local epoch ($E = 1$), i.e. a single pass over its own shard of the data, and sends the model weights to a central server for aggregation using the FedAvg~\citep{mcmahan2017communication} algorithm.

\begin{figure*}[t]
\centering
\vspace{-2mm}
\begin{overpic}[width=0.9\linewidth]{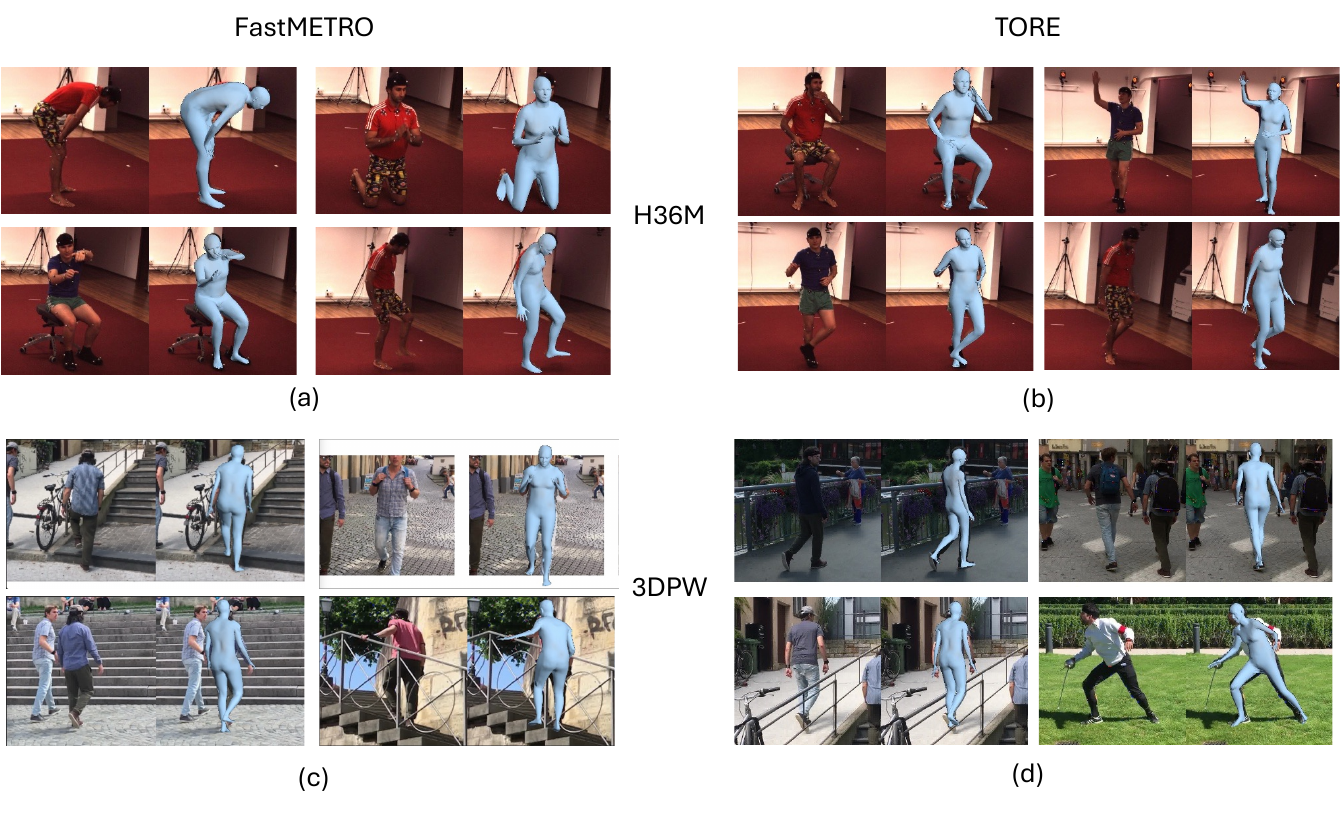}
\end{overpic}
\captionsetup{font=small,labelfont=bf}
\vspace{-2mm}
\caption{Qualitative results of \model on Human3.6M~\citep{ionescu2013human3} (a,b) and 3DPW~\citep{von2018recovering} (c,d), trained with 100 clients and 10 randomly sampled per round. The backbones are FastMETRO~\citep{cho2022cross} (a,c) and TORE~\citep{dou2023tore} (b,d).}
 \label{fig:pfhmr-visualization}
\vspace{-2mm}
\end{figure*}

As shown in Tab.~\ref{tab:basic-fl-eval}, both FastMETRO~\citep{cho2022cross} and TORE~\citep{dou2023tore} trained under the FedAvg optimizer maintain competitive accuracy even in the most challenging setting of 1000 clients, with roughly a 2--4 mm drop in MPJPE and PA-MPJPE. In the 10-client setting, the HMR models even exhibit numerically higher accuracy than their centrally trained counterparts. This underscores the robustness of the benchmarked state-of-the-art HMR models under the complex federated setting. Qualitative results for HMR models trained with the 100-client configuration are shown in Fig.~\ref{fig:pfhmr-visualization}.

While the HMR models achieve performance close to centrally trained models, one caveat is that they train much more slowly: they require many more global rounds to converge, with the most complex 1000-client setting requiring 6000 rounds. Please refer to Appendix~\ref{sec:experiment_details} for more details, including the terminology we use to describe the training budget of each regime.

\subsubsection{Performance on Skewed Data Distribution}
\label{sec:skewed-data-distribution}

In the real world, user data from user devices may vary in data distribution and sample size due to differences in identity, environment, and habits. Therefore, we manually skew the data distribution in various configurations to simulate this heterogeneity and measure its effect on HMR accuracy. We investigate three types of heterogeneous data distributions: \textbf{Latent Dirichlet Allocation (LDA)}, \textbf{Mixed Dataset}, and \textbf{Natural Partition Data}. More details on the experiment setup are in Appendix~\ref{app:skewed-data}.

\begin{figure}[b]
\centering
\begin{overpic}[width=0.6\linewidth]{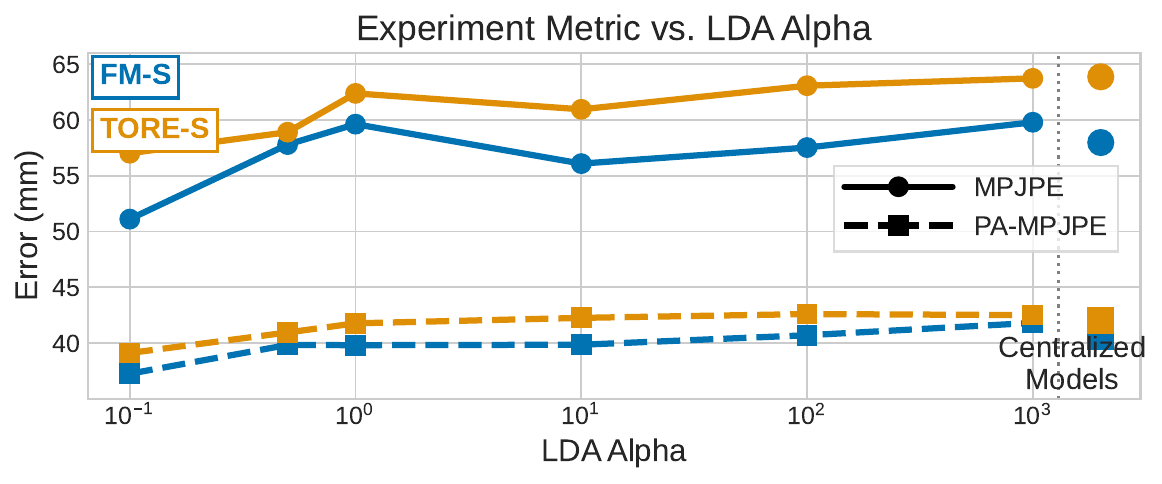}
\end{overpic}
\captionsetup{font=small,labelfont=bf}
\vspace{-3mm}
\caption{\textbf{Data Distribution Evaluation. } Performance metrics for heterogeneous data distribution settings with different LDA $\alpha$, evaluated with the Human3.6M~\citep{ionescu2013human3} dataset. }
 \label{fig:data-dist-fl-table}
\end{figure}

\noindent
\textbf{Latent Dirichlet Allocation}. As shown in \cref{fig:data-dist-fl-table}, the models under a highly heterogeneous LDA setup perform the best under the most skewed distribution, $\alpha=0.1$, with both MPJPE and PA-MPJPE errors lower than the centralized baseline for both configurations.

A plausible explanation is that local updates from highly non-uniform clients introduce an aggregation-induced regularization effect akin to meta-learning~\citep{nichol2018first}, but our current experiments do not isolate this hypothesis from centralized-run variance or remaining training-budget differences.
In addition, across a wide range of $\alpha$, the model performance is consistently competitive compared to the centralized baseline.
These results support the robustness of the benchmarked HMR models~\citep{cho2022cross,dou2023tore} under highly heterogeneous setups close to real-world scenarios.

\noindent

\begin{table}[t]
\centering
\vspace{-1mm}
\small
\resizebox{0.62\linewidth}{!}{
\begin{tabular}{lccc|c}
\toprule
\textbf{Setting} & \textbf{Model} & \textbf{MPJPE} & \textbf{PA-MPJPE} & \textbf{Raw Data Local?} \\
\midrule
Centralized & FM-S & 61.38 & 41.63 &\xmark\\
\model & FM-S & 58.28 & 40.42 &\cmark\\
\midrule
Centralized & TORE-S & 63.89 & 41.40 &\xmark\\
\model & TORE-S & 61.59 & 41.94 &\cmark\\
\bottomrule
\end{tabular}
}

\captionsetup{font=small,labelfont=bf}
\caption{\textbf{Mixed Dataset Training.} Comparison between FastMETRO-S~\citep{cho2022cross} and TORE-S~\citep{dou2023tore} trained under centralized or privacy-constrained settings on a mixed dataset composed of Human3.6M, UP-3D, COCO, MuCo-3DHP, and MPII~\citep{ionescu2013human3, lassner2017unite, lin2014microsoft, mehta2018single, andriluka20142d}, evaluated on Human3.6M.}
\label{tab:mixed-dataset-table}
\vspace{-5mm}
\end{table}

\noindent
\textbf{Mixed Dataset}. As shown in Tab.~\ref{tab:mixed-dataset-table}, HMR models trained with \model on the mixed dataset match their centralized counterparts on the majority dataset (Human3.6M): MPJPE improves by 3.1 mm for FastMETRO-S and 2.3 mm for TORE-S, and PA-MPJPE stays within 1.2 mm of the centralized result.
This suggests that HMR models trained under the privacy-constrained federated training framework can achieve competitive performance on the overall data distribution compared to the traditional centralized training framework.

\begin{table}[t]
\centering
\small
\resizebox{0.7\linewidth}{!}{
\begin{tabular}{lcccc|c}
\toprule
\textbf{Setting} & \textbf{Model} & \textbf{MPJPE} & \textbf{PA-MPJPE} & \textbf{MPVPE} & \textbf{Raw Data Local?} \\
\midrule
Centralized & FM-S & 84.92 & 54.69 & 97.60 &\xmark\\ \midrule
\model & FM-S & 84.66 & 54.78 & 97.42 &\cmark\\
\model (Natural Part.) & FM-S & 86.83 & 55.61 & 99.66 & \cmark\\
\midrule
Centralized & TORE-S & 87.97 & 55.35 & 101.88 &\xmark\\ \midrule
\model & TORE-S & 87.55 & 54.08 & 101.86 & \cmark\\
\model (Natural Part.) & TORE-S & 88.00 & 55.07 & 102.50 & \cmark\\
\bottomrule
\end{tabular}
}

\captionsetup{font=small,labelfont=bf}
\caption{\textbf{Performance of \model on 3DPW.} Reconstruction accuracy when \model is fine-tuned on the 3DPW dataset. We further evaluate the model's performance on the natural scene partition (Natural Part.), where each scene is assigned to a device. \model reaches comparable performance on 3DPW fine-tuning compared with the centralized setting. \model is trained with 100 devices and 10 randomly selected each round, while the natural partition is 20 devices with 5 randomly selected each round.}
\label{tab:pre-train-fl-3dpw-finetune-table}
\vspace{-2mm}
\end{table}

\noindent
\textbf{Natural Partition Data}. As shown in Tab.~\ref{tab:pre-train-fl-3dpw-finetune-table}, under both partition schemes, FastMETRO and TORE stay within 2.1 mm of their centralized counterparts on every metric; the largest gap is FastMETRO under the natural partition (+1.9 mm MPJPE, +2.1 mm MPVPE). The results show that privacy-constrained distributed fine-tuning of HMR models can achieve performance comparable to centralized fine-tuning, even under the challenging natural partition scheme.

\begin{table}[h!]
\centering
\small
\captionsetup{labelfont=bf}
\resizebox{0.7\linewidth}{!}{%
\begin{tabular}{lccccc}
\toprule
\textbf{Dataset} & \textbf{Before/After} & \textbf{Model} & \textbf{MPVPE} & \textbf{MPJPE} & \textbf{PA-MPJPE}\\
\midrule
VR-runner & Before \model & FM-S & 109.50 & 114.51 & 60.07 \\
VR-runner & After \model & FM-S & 93.23 & 58.24 & 43.16 \\
\midrule
VR-game-1 & Before \model & FM-S & 176.24 & 172.57 & 78.21 \\
VR-game-1 & After \model & FM-S & 65.57 & 62.30 & 43.82 \\
\midrule
VR-game-2 & Before \model & FM-S & 150.64 & 161.82 & 84.36 \\
VR-game-2 & After \model & FM-S & 82.53 & 69.49 & 53.57 \\
\midrule
Oculus & Before \model & FM-S & 103.92 & 102.22 & 61.62 \\
Oculus & After \model & FM-S & 56.76 & 57.92 & 39.94 \\
\bottomrule
\end{tabular}
}
\captionsetup{font=small,labelfont=bf}
\caption{\textbf{Local Fine-Tuning Results.} Reconstruction error before and after local fine-tuning with \model, computed on the held-out validation split of the locally annotated data against DePoser pseudo-labels (Appendix~\ref{sec:local_annotation}).}

\vspace{-5mm}
\label{tab:local_finetuning}
\end{table}

\subsubsection{Comparison between Federated Optimizers}

In the previous section, experiment results confirmed the robustness of FastMETRO~\citep{cho2022cross} and TORE~\citep{dou2023tore} under different forms of data skewness. We now benchmark the HMR models on three popular federated learning algorithms, namely FedAvg~\citep{mcmahan2017communication}, FedOpt~\citep{reddi2021adaptive}, and FedProx~\citep{li2020federated}.
The latter two more sophisticated federated optimizers offer better adaptiveness and robustness than FedAvg.

As shown in Tab.~\ref{tab:fed-optimizers}, in most of the settings evaluated, the three optimizers performed similarly in terms of both MPJPE and PA-MPJPE, with no clear trend of {superiority or inferiority}. This suggests the robustness of HMR models coupled with the simplest FedAvg~\citep{mcmahan2017communication} under a highly skewed data setting (LDA $\alpha=0.5$).

\subsubsection{Personalized Fine-tuning}

Although HMR models could maintain competitive performance under challenging real-world multi-device collaborative training settings, further personalized fine-tuning is often desired to tailor the model to a specific user's data distribution.
Therefore, we benchmark further fine-tuning a collaboratively pre-trained FastMETRO~\citep{cho2022cross} model on additional in-the-wild datasets that reflect real-world use cases, such as VR gaming and sports. We use our local data annotator, DePoser, to annotate pseudo-ground truth for the collected dataset. For local fine-tuning, we divide each scene of our dataset into an 80/20 random split and fine-tune the pre-trained model for two epochs over the user's local training split.

\begin{figure}[t]
\centering

\begin{minipage}[t]{0.34\linewidth}
\vspace{0pt}
\centering
\begin{overpic}[width=\linewidth]{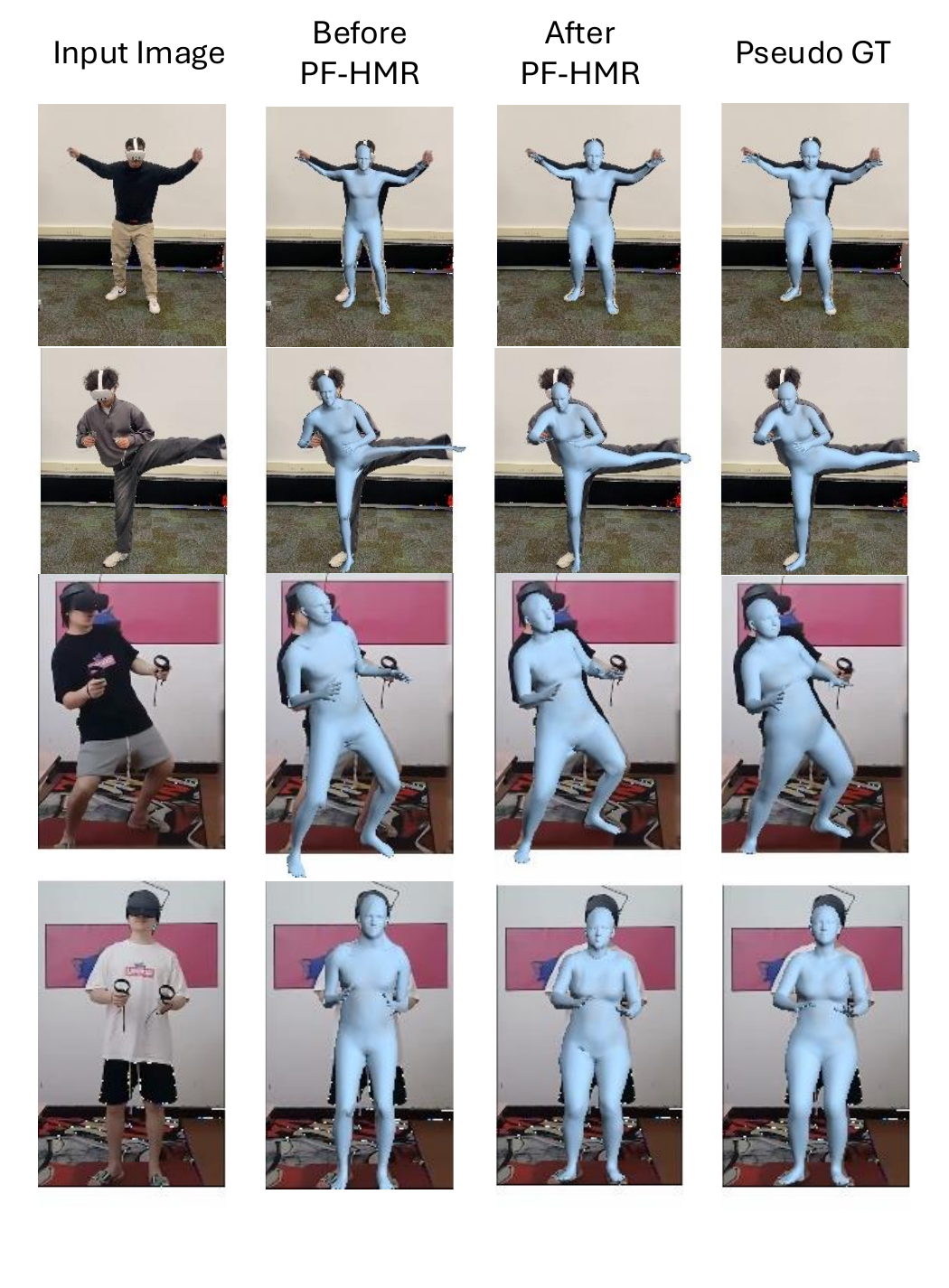}
\end{overpic}
\captionsetup{font=small,labelfont=bf}
\caption{Qualitative comparison of FastMETRO~\citep{cho2022cross} before and after \model fine-tuning on a locally annotated dataset, with pseudo-ground truth (GT) generated by DePoser.}
\label{fig:local_finetuning}
\end{minipage}
\hfill
\begin{minipage}[t]{0.64\linewidth}
\vspace{0pt}
\centering
\begin{overpic}[width=\linewidth]{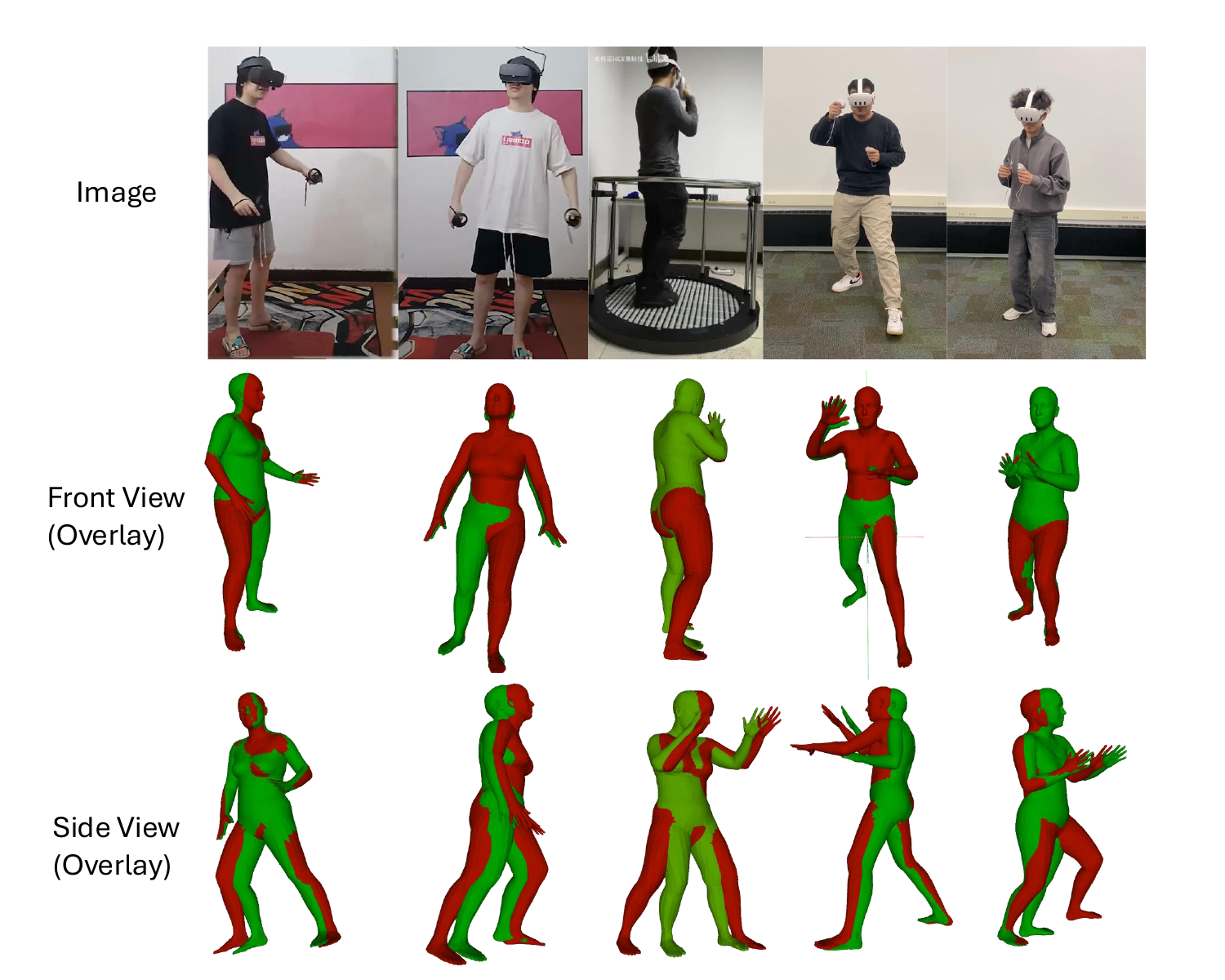}
\end{overpic}
\captionsetup{font=small,labelfont=bf}
\caption{Comparison between our DePoser (green) and SMPLify-X (red)~\citep{pavlakos2019expressive}. DePoser exhibits more accurate foot placing and hand positioning under the combined Sapiens pose and depth prior.}
\label{fig:depth-ambiguity}
\end{minipage}

\end{figure}

As shown in Fig.~\ref{fig:local_finetuning}, our personalized fine-tuned model reconstructs human meshes that more accurately resemble the human pose and shape after fine-tuning on the pseudo-GT. Quantitatively, as shown in Tab.~\ref{tab:local_finetuning}, MPVPE drops by up to 110.7 mm after local fine-tuning, with the largest improvement in VR-game-1, a challenging home VR gaming setting. Because the evaluation labels in this table are generated by DePoser, these numbers measure consistency with the local pseudo-label distribution rather than accuracy against independent ground truth. In addition, we perform the personalized fine-tuning using the 3DPW dataset and observe a similar performance improvement. Details are shown in Appendix~\ref{app:3dpw-finetune}. This demonstrates that the strong prior of an HMR model pre-trained on a large amount of data can enable effective personalization with only two epochs of fine-tuning, significantly improving end-user performance through model personalization.

\begin{table}[htbp]
\centering
\small
\captionsetup{labelfont=bf}

\resizebox{0.5\linewidth}{!}{%
\begin{tabular}{l|ccc}
\toprule
\textbf{Model} & \textbf{MPVPE} & \textbf{MPJPE} & \textbf{PA-MPJPE} \\
\midrule
SMPLify-X~\citep{pavlakos2019expressive} & 198.66 & 181.07 & 87.17 \\
DePoser (Ours) & \textbf{162.60} & \textbf{151.19} & \textbf{71.14} \\
\bottomrule
\end{tabular}
}

\captionsetup{font=small,labelfont=bf}
\caption{\textbf{Comparison between DePoser and SMPLify-X~\citep{pavlakos2019expressive}} on a sub-sampled 3DPW~\citep{von2018recovering} dataset. DePoser demonstrates improved reconstruction accuracy compared to SMPLify-X under the combined Sapiens pose and depth prior.}

\label{tab:deposer-eval}
\end{table}

\subsection{Local Annotation Pipeline}
\label{sec:local-annotation-eval}

Privacy-constrained HMR training depends on accurate local annotation. We therefore compare DePoser with SMPLify-X~\citep{pavlakos2019expressive} to measure the effect of the Sapiens~\citep{khirodkar2024sapiens} pose and depth priors. The comparison evaluates the full DePoser pipeline and does not ablate its components (keypoint source, depth term, metric-scale alignment).
More details are available in Appendix~\ref{sec:local_annotation}.

\textbf{Quantitative Analysis.} We compare the human reconstruction accuracy of our DePoser and SMPLify-X~\citep{pavlakos2019expressive} on a 300-image randomly sub-sampled set of the in-the-wild dataset 3DPW~\citep{von2018recovering}, which resembles complex user-side conditions. As shown in Tab.~\ref{tab:deposer-eval}, DePoser demonstrates improved accuracy compared to SMPLify-X, with more than 30 mm accuracy improvement in MPVPE and 15 mm improvement in PA-MPJPE, supporting the effectiveness of the combined local annotation pipeline under complex conditions.

\textbf{Qualitative Analysis.} As shown in Fig.~\ref{fig:depth-ambiguity}, although the front-view results of DePoser and SMPLify-X~\citep{pavlakos2019expressive} are similar, DePoser places the feet and arms more accurately along the z-axis (depth) than SMPLify-X, which is evident in the side view. This visual evidence is consistent with the quantitative improvement. The figure compares two annotators that differ in keypoint source, depth term, and depth alignment simultaneously.

\vspace{-2mm}

\section{Limitations and Future Work}

\noindent
\textbf{HMR-aware Aggregation Strategy.} Our LDA results suggest that aggregating updates from heterogeneous clients may regularize HMR training. We have not verified this mechanism; matched-budget controlled studies are needed to separate it from optimization variance and schedule differences.

\noindent
\textbf{Device-specific and Personalized Learning.} In real-world human mesh recovery, each device would introduce unique data perturbation into the client data distribution, such as smearing, tearing, and fuzziness. However, there is no existing solution for simulating this issue. Having a personalized model adapted to device perturbations would be a promising direction for future work.

\noindent
\textbf{Privacy Evaluation Scope.} Our experiments measure HMR utility with raw images kept local and the runtime cost of a secure-aggregation variant. We do not evaluate privacy attacks or provide formal guarantees (see the Broader Impact Statement); both are important future work.

\noindent
\textbf{Local Annotation Bias.} DePoser produces SMPL-parametric pseudo-annotations, which are then used to supervise non-parametric HMR regressors. The personalization gains therefore measure agreement with DePoser's pseudo-labels and are not validated against independent ground truth. The DePoser comparison also evaluates the full pipeline (Sapiens pose, Sapiens depth, metric alignment) without ablating each component.

\vspace{-2mm}

\section{Conclusion}
In this work, we introduced \model, a benchmark for evaluating human mesh recovery (HMR) models under privacy-constrained, raw-data-local training settings. Across client scales and data-heterogeneity settings, we showed that state-of-the-art HMR models can be collaboratively trained with a large number of clients and heterogeneous data while incurring only minimal accuracy loss. Our study also reveals remaining challenges, including slower convergence and degraded performance when generalizing to out-of-distribution data, which mark remaining gaps between centralized and privacy-constrained training regimes. To address the data availability bottleneck in privacy-constrained HMR, we proposed \emph{DePoser}, a local data annotation pipeline that conditions not only on 2D keypoints but also on pseudo-depth information. Comparisons against SMPLify-X demonstrate that DePoser improves pseudo-annotation quality under the combined Sapiens pose and depth priors. Using DePoser, we curated and annotated an in-the-wild dataset of VR use cases, and showed that privacy-constrained fine-tuning on this dataset yields substantial gains for HMR model personalization on end-user devices when evaluated against the same pseudo-label source. We hope that our benchmark, analysis, and the DePoser pipeline will serve as a foundation for future research on privacy-constrained human mesh recovery, including more robust collaborative training algorithms, stronger privacy guarantees, and better on-device personalization strategies.

\section*{Broader Impact Statement}

This work trains and evaluates human mesh recovery models on images of people. Two aspects therefore warrant explicit discussion: the provenance of the in-the-wild dataset we collected for the personalization study, and how the term \textit{privacy-constrained}, used throughout the paper, should and should not be read.

\noindent \textbf{Data Provenance and Consent.} The benchmark experiments use only established public HMR datasets~\citep{ionescu2013human3, lassner2017unite, lin2014microsoft, andriluka20142d, mehta2018single, von2018recovering} under their original terms, none of which we redistribute. The VR dataset used for personalization (\cref{sec:experiment-setup}) is new and has two subsets. The in-house subset was recorded from participants who gave informed consent to the research use of their footage. This footage is used for training and evaluation only and is not redistributed. The internet subset is drawn from publicly accessible videos of people using consumer VR and AR devices. These individuals are identifiable and were not asked for consent to be included in a research dataset, which we regard as the most sensitive aspect of this work. We use the footage solely to fit and evaluate body meshes: no face recognition, identity linkage, or demographic attribute inference is performed, and none is released. With four scenes and four identities, the dataset is a personalization stress test rather than a demographic coverage benchmark, and models fine-tuned on it should not be assumed to behave equitably across body shapes, skin tones, or mobility profiles.

\noindent \textbf{Release Plan.} We release code, not data (\cref{sec:experiment-setup}). Upon publication we will make the \model harness and the DePoser implementation available, covering the training, evaluation, and data-splitting code for both, so that our experiments can be reproduced end to end. We redistribute no imagery. The public HMR datasets we train and evaluate on remain the property of their respective rights holders and must be obtained from them under their original terms; for these we provide a guide to preparing the data and replicating our results rather than a copy of the data. The VR footage curated for the personalization study is not released in any form, neither the internet-sourced videos nor the pseudo-annotations derived from them. This is the most conservative option open to us, and we prefer it to the alternative: it costs reproducibility, since the personalization experiments cannot be re-run from our release without collecting comparable footage, but it avoids redistributing imagery of identifiable people who never consented to research use, and avoids attaching body-mesh annotations to those people in a public artifact.

\noindent \textbf{The Limits of ``Privacy-Constrained''.} We use \textit{privacy-constrained} to name a training setting, one in which raw images and mesh annotations never leave the client device, and not to assert a privacy guarantee. Raw-data locality is not privacy: individual model updates, and the released global model, may still leak information about the local data that produced them~\citep{kairouz2021advances}. We perform no attack-based privacy evaluation in this work, and therefore do not claim protection against membership-inference, gradient-inversion, or model-inversion attacks. The secure-aggregation variant assumes an honest-but-curious server and prevents it from observing individual client updates, but we report it as a small-scale runtime feasibility check rather than a security analysis, and we do not claim formal differential privacy for any model we release. Our results should accordingly not be read as establishing regulatory compliance, nor as warrant for training HMR models on user data without the additional safeguards, such as formal guarantees and empirical attack-based audits, that a real deployment would require.

\bibliography{main}
\bibliographystyle{tmlr}

\clearpage
\appendix

This supplementary material covers: More Experiment Details~(\cref{sec:experiment_details}); Training Budget of the Centralized and Federated Regimes~(\cref{app:training-budget}); More Details on Skewed Data Distribution~(\cref{app:skewed-data}); Local Data Annotation~(\cref{sec:local_annotation}); More Implementation Details~(\cref{sec:imp_details}); Discussion on MPC and DP~(\cref{app:mpc_dp}); Discussion on Single-Device Training~(\cref{app:naive-single-device-training}); Discussion on Multiple Local Epochs~(\cref{app:local_epoch}); Discussion on Performance of 3DPW Fine-tuning~(\cref{app:3dpw-finetune}); Discussion on Data Distribution~(\cref{app:data_distribution}); Additional Experiments for On-Device Performance~(\cref{app:on-device-perf}).

\section{Experiment Hyperparameters}
\label{sec:experiment_details}

\noindent \textbf{Terminology.} We follow the standard federated-learning vocabulary~\citep{mcmahan2017communication, bonawitz2019towards, kairouz2021advances} and reserve each term for exactly one quantity. A \textit{global round} (or simply \textit{round}) is one server-side aggregation cycle: the server samples a set of clients, each of them trains on its own data, and the resulting model updates are aggregated into a new global model; $T$ denotes the number of rounds. $K$ is the total number of clients and $m$ the number sampled uniformly at random per round, reported throughout as \texttt{$K$/$m$}. A \textit{local epoch} ($E$) is one pass a sampled client makes over its own shard of $n_i$ samples within a round. A \textit{local step}, equivalently an \textit{optimizer step}, is a single gradient update on one mini-batch, so a client performs $\tau_i = E \lceil n_i / B \rceil$ of them per round at local batch size $B$. A \textit{client update} is the payload a client returns to the server, not a gradient update. The unqualified word \textit{epoch} always means one pass over the \textit{full} training set and is used only for centralized training; federated budgets are quoted in rounds. To compare the two regimes on a common axis we count \textit{sample-visits} (one image passed through the model once) and convert them to \textit{epoch-equivalents} by dividing by the size of the training set.

Fixing $E$ fixes the number of local passes but not the amount of local work, because $\tau_i$ grows with the shard size $n_i$. Under the IID partitions all shards are the same size and the distinction is immaterial, but under the skewed partitions of~\cref{app:skewed-data} the shards differ by orders of magnitude, so a single local epoch amounts to very different numbers of optimizer steps across clients. We therefore report both quantities where the distinction matters.

In Tab.~\ref{tab:hyperparameters} we present the hyperparameters for the experiments in Sec.~\ref{sec:evaluation} of the main paper. The \textit{Epochs} column applies only to the centralized runs, which have no rounds; the federated runs are instead specified by their local epochs $E$ and their global rounds $T$.

\begin{table}[htb]
\centering
\small
\resizebox{0.95\columnwidth}{!}{%
\begin{tabular}{lcccccc}
\toprule
\textbf{Setting} & \textbf{Model} & \textbf{Clients } & \textbf{Learning Rate} & \textbf{Epochs} & \textbf{Local Epochs ($E$)} & \textbf{Global Rounds ($T$)} \\
\midrule
Centralized Baseline & FM-S & - & 0.001 & 60 & -- & -- \\
Human3.6M na\"ive on-device training & FM-S & - & 0.001 & 60 & -- & -- \\
3DPW Centralized Fine-tuning & FM-S & - & 0.0001 & 20 & -- & -- \\
\midrule
\model + IID Data 10 / 10 & FM-S & 10 / 10 & 0.001 & -- & 1 & 200 \\
\model + IID Data 100 / 10 & FM-S & 100 / 10 & 0.001 & -- & 1 & 2000 \\
\model + IID Data 1000 / 10 & FM-S & 1000 / 10 & 0.001 & -- & 1 & 6000 \\
\model + LDA Distribution & FM-S & 100 / 10 & 0.001 & -- & 1 & 2000 \\
\model + CNN Cluster & FM-S & 100 / 10 & 0.001 & -- & 1 & 2000 \\
\model + 3DPW Fine-tuning & FM-S & 100 / 10 & 0.0001 & -- & 1 & 600 \\
\model + 3DPW Natural Part. & FM-S & 20 / 5 & 0.0001 & -- & 1 & 200 \\

\midrule
Centralized Baseline & TORE-S & - & 0.001 & 60 & -- & -- \\
Human3.6M na\"ive on-device training & TORE-S & - & 0.001 & 60 & -- & -- \\
3DPW Centralized Fine-tuning & TORE-S & - & 0.0001 & 20 & -- & -- \\
\midrule
\model + IID Data 10 / 10 & TORE-S & 10 / 10 & 0.0005 & -- & 1 & 200 \\
\model + IID Data 100 / 10 & TORE-S & 100 / 10 & 0.0005 & -- & 1 & 2000 \\
\model + IID Data 1000 / 10 & TORE-S & 1000 / 10 & 0.001 & -- & 1 & 6000 \\
\model + LDA Distribution & TORE-S& 100 / 10 & 0.0005 & -- & 1 & 2000 \\
\model + CNN Cluster & TORE-S & 100 / 10 & 0.0005 & -- & 1 & 2000 \\
\model + 3DPW Fine-tuning & TORE-S & 100 / 10 & 0.0001 & -- & 1 & 600 \\
\model + 3DPW Natural Part. & TORE-S & 20 / 5 & 0.0001 & -- & 1 & 200 \\

\bottomrule
\end{tabular}
}
\captionsetup{font=small,labelfont=bf}
\caption{The hyperparameters used by all \model experiments. The two numbers in the Clients column denote the total number of clients and the number of clients randomly selected in each round. \textit{Epochs} counts passes over the full training set and applies only to the centralized runs; the federated runs are specified instead by the number of local epochs $E$ each sampled client performs per round and the number of global rounds $T$. The two units are not interchangeable, and we quote a budget in only one of them per run.}
\label{tab:hyperparameters}
\vspace{-3mm}
\end{table}

\section{Training Budget of the Centralized and Federated Regimes}
\label{app:training-budget}

Because several of our comparisons place a federated run next to a centrally trained baseline, we report here, for every regime in the paper, the quantities that determine how much training each of them actually received: total optimizer steps, total data passes, wall-clock cost, checkpoint-selection rule and validation protocol. \Cref{tab:training-budget} collects them.

The quantities in \cref{tab:training-budget} follow from a few fixed properties of our setup. The training corpus contains $476{,}062$ images, so one epoch-equivalent is $476{,}062$ sample-visits. Every federated client trains with a local batch of $B=64$ on a single device, whereas the centralized baselines are trained with data parallelism over eight devices and therefore have an \textit{effective} batch of $512$, i.e. $\lceil 476{,}062/512 \rceil = 930$ optimizer steps per epoch. This asymmetry is the reason the two regimes cannot be matched in data passes and in gradient steps by the same run, and it is why we also list a hypothetical single-device centralized run at $B=64$. For the federated rows we additionally report the \textit{critical path}: the number of optimizer steps summed over rounds of the slowest client sampled in each round, which is the update depth a deployment would actually wait through, since a round cannot close until its last client reports. Total steps and sample-visits are backbone-independent and are obtained from the partition sizes and the sampling rule; wall-clock is the one quantity that depends on the backbone, so we report FastMETRO-S and TORE-S separately. The step-matched configuration, which we did not run to completion, is extrapolated from 200-epoch runs.

\begin{table}[htbp]
\centering
\small
\captionsetup{labelfont=bf}
\resizebox{\columnwidth}{!}{%
\begin{tabular}{llcrrrrrrr}
\toprule
 & & & & & & & \multicolumn{2}{c}{\textbf{Wall-clock (h)}} & \\
\cmidrule(lr){8-9}
\textbf{Regime} & \textbf{Budget} & \textbf{Eff.\ batch} & \textbf{Optimizer steps} & \textbf{Sample-visits} & \textbf{Epoch-equiv.} & \textbf{Critical path} & \textbf{FM-S} & \textbf{TORE-S} & \textbf{Val.\ evals} \\
\midrule
Centralized (as published) & 60 epochs & 512 & $55{,}800$ & $28{,}563{,}720$ & $60.0$ & -- & $18.9$ & $18.7$ & $60$ \\
Centralized, budget-matched & 200 epochs & 512 & $186{,}000$ & $95{,}212{,}400$ & $200.0$ & -- & $63.0$ & $62.3$ & $200$ \\
Centralized, step-matched & 200 epochs & 64 & $1{,}487{,}800$ & $95{,}212{,}400$ & $200.0$ & -- & $\approx 500^{\dagger}$ & $\approx 500^{\dagger}$ & $200$ \\
\midrule
\model, IID \texttt{10/10} & 200 rounds & 64 & $1{,}488{,}000$ & $95{,}212{,}400$ & $200.0$ & $148{,}800$ & $44.3$ & $47.1$ & $200$ \\
\model, IID \texttt{100/10} & 2000 rounds & 64 & $1{,}500{,}203$ & $95{,}212{,}580$ & $200.0$ & $150{,}203$ & $60.5$ & $55.9$ & $200$ \\
\model, IID \texttt{1000/10} & 6000 rounds & 64 & $480{,}060$ & $28{,}563{,}726$ & $60.0$ & $48{,}060$ & $56.3$ & $60.6$ & $600$ \\
\midrule
\model, LDA $\alpha=0.1$ & 2000 rounds & 64 & $1{,}496{,}597$ & $95{,}124{,}910$ & $199.8$ & $707{,}920$ & $87.4$ & $77.7$ & $200$ \\
\model, LDA $\alpha=0.5$ & 2000 rounds & 64 & $1{,}496{,}536$ & $95{,}151{,}465$ & $199.9$ & $437{,}122$ & $65.2$ & $61.5$ & $200$ \\
\model, LDA $\alpha=1$ & 2000 rounds & 64 & $1{,}497{,}281$ & $95{,}207{,}630$ & $200.0$ & $347{,}585$ & $60.9$ & $59.0$ & $200$ \\
\model, LDA $\alpha=10$ & 2000 rounds & 64 & $1{,}497{,}717$ & $95{,}228{,}649$ & $200.0$ & $204{,}363$ & $55.5$ & $56.0$ & $200$ \\
\model, LDA $\alpha=100$ & 2000 rounds & 64 & $1{,}497{,}416$ & $95{,}215{,}663$ & $200.0$ & $166{,}057$ & $59.4$ & $57.6$ & $200$ \\
\model, LDA $\alpha=1000$ & 2000 rounds & 64 & $1{,}497{,}387$ & $95{,}213{,}672$ & $200.0$ & $154{,}767$ & $59.2$ & $57.4$ & $200$ \\
\bottomrule
\end{tabular}
}
\captionsetup{font=small,labelfont=bf}
\caption{\textbf{Training Budget of Every Regime.} Optimizer steps are gradient updates summed over all clients; sample-visits count one image passed through the model once, and epoch-equivalents divide them by the $476{,}062$-image training set; the critical path counts the steps of the slowest client sampled in each round, summed over rounds. All federated rows use one local epoch ($E=1$) and a local batch of $64$; the LDA rows use $100$ clients with $10$ sampled per round and are averaged over partition and client-sampling seeds. Wall-clock is the only backbone-dependent quantity and is reported separately for each; measured values are the median over the runs of that backbone that completed their configured budget in that setting, and $^{\dagger}$ marks a projected value for the step-matched configuration. Runs were executed on a shared cluster with heterogeneous nodes, so wall-clock should be read as an order-of-magnitude comparison rather than a controlled timing benchmark, and it measures our simulation cost rather than the latency of a real deployment; \textit{Val.\ evals} is the number of validation passes performed over a run.}
\label{tab:training-budget}
\end{table}

\noindent \textbf{Validation protocol and checkpoint selection.} Both regimes are validated on exactly the same data, the Human3.6M protocol-2 split of $27{,}558$ images, with a validation batch of $256$, and the reported metrics are computed on that split as well, so no separate held-out test set is used; the numbers should be read accordingly. Federated evaluation is server-side and centralized: the aggregated global model, not any client model, is evaluated on the full validation set, and client-side validation is disabled. Centralized runs evaluate and checkpoint once per epoch; federated runs evaluate and checkpoint once per round for the 200-round setting and once every ten rounds for the 2000- and 6000-round settings, which yields the evaluation counts in the last column of \cref{tab:training-budget}. In every regime the value we report is the best evaluation on that validation set over the course of the run, i.e. best epoch for centralized runs and best round for federated ones.

\noindent \textbf{Centralized versus federated.} The comparison is not budget-matched, and the mismatch goes in opposite directions depending on which quantity one counts. In data passes, the \texttt{10/10}, \texttt{100/10} and all LDA runs receive $200$ epoch-equivalents against the published baseline's $60$, a factor of $3.33$; the \texttt{1000/10} run is the only configuration matched in data passes, and it is also the configuration that trails the centralized baseline on every metric. In gradient steps the ordering reverses: the baseline takes only $55{,}800$ optimizer steps because data parallelism raises its effective batch to $512$, whereas the federated clients train at batch $64$ and take between $480{,}060$ and $1{,}500{,}203$, so even the data-matched \texttt{1000/10} run performs $8.6\times$ more updates than the baseline it is compared against. No single centralized re-run can close both gaps at once unless the batch size is matched too: a $200$-epoch run at batch $512$ equalizes sample-visits only, while a $200$-epoch run at batch $64$ equalizes both, at $1{,}487{,}800$ steps against \texttt{10/10}'s $1{,}488{,}000$ on identical sample-visits. Wall-clock is not matched either, and not in the federated regime's favor: the published baseline costs $18.7$\,h for TORE-S and about $18.9$\,h for FastMETRO-S, against $44.3$--$87.4$\,h for the federated runs, so every federated configuration is between two and five times as expensive to train as the baseline it is compared against, on both backbones. We therefore do not attribute any accuracy the federated runs gain over the centralized baseline solely to the federated protocol itself: the comparisons in which \model matches or slightly exceeds centralized training were obtained with more data passes and far more gradient updates than the baseline received, and the one configuration that is matched in data passes (\texttt{1000/10}) does not win in accuracy.

\begin{figure}[htbp]
 \centering
 \begin{subfigure}[b]{\columnwidth}
 \centering
 \includegraphics[width=\textwidth]{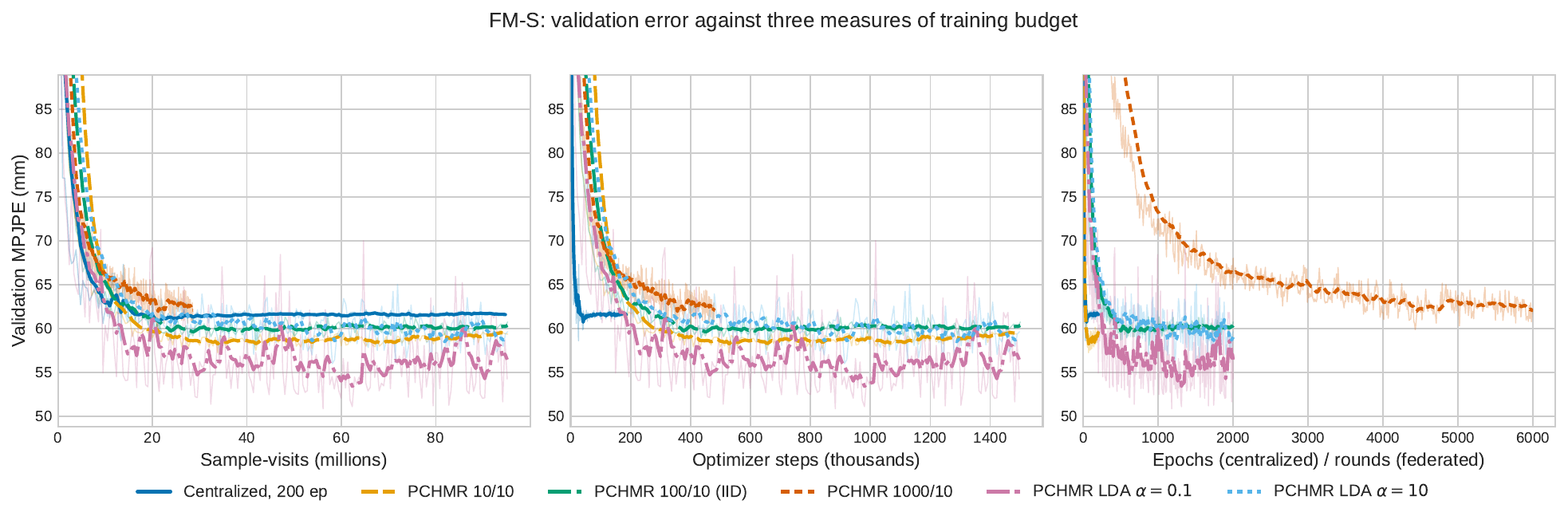}
 \caption{FastMETRO-S}
 \label{fig:budget-curves-fm}
 \end{subfigure}
 \\[0.6em]
 \begin{subfigure}[b]{\columnwidth}
 \centering
 \includegraphics[width=\textwidth]{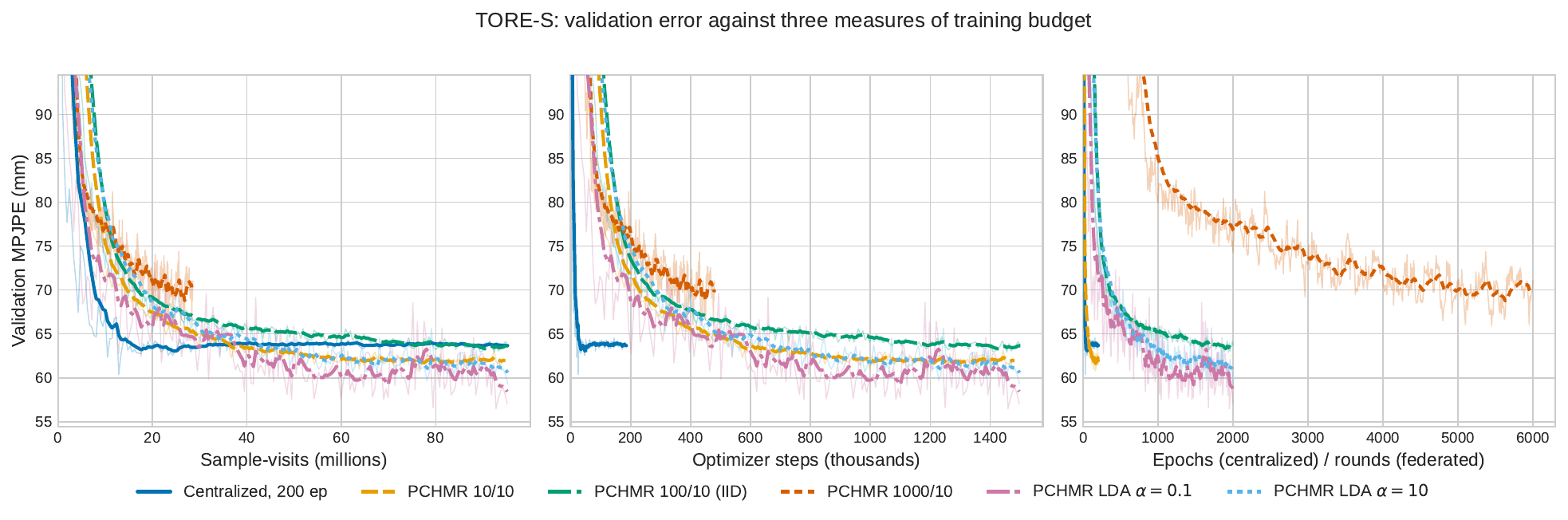}
 \caption{TORE-S}
 \label{fig:budget-curves-tore}
 \end{subfigure}
\captionsetup{font=small,labelfont=bf}
\caption{\textbf{Validation Error against Three Measures of Training Budget.} Each row shows the same six runs for one backbone --- the budget-matched $200$-epoch centralized re-run, the three IID client scales, and the LDA partitions at $\alpha=0.1$ and $\alpha=10$ --- plotted against sample-visits (left), optimizer steps (middle), and the native progress counter each regime was logged on, epochs centrally and communication rounds federated (right). The left two panels place the two regimes on a common axis and are the comparison \cref{tab:training-budget} quantifies; the right panel is what that comparison looks like when it is \emph{not} aligned, and is included because epochs and rounds are not interchangeable units. Faint traces are the raw per-epoch and per-round evaluations and bold traces an exponential moving average over them. All axes are linear, and the vertical range is clipped to the region where the curves separate, so the first few evaluations of each run fall above the panel. On the right-hand panel the horizontal extent of a curve is set by the number of rounds its configuration was run for rather than by the training it received, which is why \texttt{1000/10} spans six times the axis of the runs it is compared against while receiving less than a third of their gradient updates. }
\label{fig:budget-curves}
\end{figure}

\noindent \textbf{Learning curves.} \Cref{fig:budget-curves} plots the validation error of the budget-matched centralized re-run and the federated runs against each of these budget measures. Read against sample-visits, the two regimes are directly comparable, and the $200$-epoch centralized run reaches the same number of visits as the \texttt{10/10}, \texttt{100/10} and LDA runs. Read against optimizer steps, their positions on the axis reverse, since the centralized run covers the same data in $8$--$27\times$ fewer, larger steps. Three features of the curves matter for the comparison. First, the centralized runs reach their best validation error early --- epoch $26$ for TORE-S and epoch $27$ for FastMETRO-S, both well inside the published $60$-epoch budget --- and then flatten and slowly degrade, so the centralized baseline is not under-trained and the extra data passes are spent past the point of convergence. This is the convergence-matched reading of the comparison, and it holds independently of the sample-visit matching. Second, the federated curves are still descending where the centralized curve has flattened, so the two regimes are at different points of their trajectories at the moment they are compared. Third, the federated curves are visibly noisier from evaluation to evaluation, because each round evaluates a model aggregated from a different sample of clients, and that noise grows as the partition becomes more skewed.

\noindent \textbf{The effect of $\alpha$ on the budget.} Changing the Dirichlet concentration does not change how much training the model receives. Across $\alpha \in [0.1, 1000]$ the LDA runs all take $\approx\!1.497$M optimizer steps and see $199.8$--$200.0$ centralized epoch-equivalents, within $0.5\%$ of the IID \texttt{100/10} run at the same number of rounds. This is structural rather than coincidental: the Dirichlet partition assigns all $476{,}062$ images regardless of $\alpha$, and clients are sampled uniformly, so the expected amount of data processed per round is the same at every skew, and only its round-to-round variance grows as $\alpha$ falls. What skew does change is how that fixed amount of work is distributed across the clients of a round, which is captured by the critical path: it grows $4.6\times$ from $154{,}767$ steps at $\alpha=1000$ to $707{,}920$ at $\alpha=0.1$, because at high skew nearly every round samples at least one client holding a very large shard and must wait for it. Wall-clock follows the critical path rather than the total step count, and does so on both backbones: it rises from $57.4$\,h at $\alpha=1000$ to $77.7$\,h at $\alpha=0.1$ for TORE-S and from about $59.2$\,h to $87.4$\,h for FastMETRO-S, increases of $35\%$ and $48\%$ for the same number of gradient updates. Relative to the IID run at the same number of rounds the penalty is of the same size, $39\%$ for TORE-S and $44\%$ for FastMETRO-S. The same straggler effect would appear, more severely, in a real cross-device deployment, where the slowest device is slow for reasons other than shard size. The practical consequence for this paper is that training budget cannot explain our most surprising observation, that the most skewed partitions perform comparably to or slightly better than centralized training: the IID \texttt{100/10} run has an identical budget and does not show the effect, so the difference between them lies in the partition and the aggregation dynamics it induces, not in the amount of training. Budget does, however, remain a live explanation for the federated-versus-centralized comparison itself, as discussed above.

\section{Skewed Data Distribution}
\label{app:skewed-data}

We present the details of the experimental setup for the three skewed data distributions discussed in~\cref{sec:skewed-data-distribution}: \textbf{Latent Dirichlet Allocation (LDA)}, \textbf{Mixed Dataset}, and \textbf{Natural Partition Data}.

\noindent
\textbf{Latent Dirichlet Allocation.} To test the HMR model's performance under a heterogeneous data setting without any prior partition information available, we follow~\citet{tang2021data, hsu2019measuring, grativol2023federated} in using Dirichlet-based label partitioning, referred to as Latent Dirichlet Allocation (LDA) in the FL literature, for generating skewed training datasets to approximate real-world data distributions. The LDA partition scheme uses the $\alpha$ parameter to control the skewness of the generated partition with the label. A smaller $\alpha$ corresponds to a more skewed distribution, while a larger $\alpha$ indicates that the distribution is close to IID (independent and identically distributed). More details on LDA are discussed in Appendix~\ref{app:lda-data-partition}. For LDA, we investigate FastMETRO and TORE's performance with 100 clients and 10 clients per round, trained with datasets of varying skewness controlled by LDA.

\noindent
\textbf{Mixed Dataset.} In addition to simulated label skewness, we also implement a mixed dataset which blends the training datasets~\citep{ionescu2013human3, lassner2017unite, lin2014microsoft, andriluka20142d, mehta2018single}. We allocate each client to a source dataset according to the dataset size ratio, which creates a heterogeneous training corpus across clients. The experiment is performed under the 100-client setting with 10 clients fitted per round.

\noindent
\textbf{Natural Partition Data.} We also present a study on 3DPW~\citep{von2018recovering}, an in-the-wild human mesh dataset collected in various indoor and outdoor scenes, which provides a natural way to partition the dataset by scenes, resulting in a highly heterogeneous per-client data distribution. Under this data partitioning, we explore the performance of the HMR model trained using a privacy-constrained method. Further information on the 3DPW natural partition is available in Appendix~\ref{app:3dpw-natural-partition}. As a baseline, we first fine-tune pre-trained HMR models on the 3DPW dataset under the federated framework using an IID partitioning scheme. Then, to explore the effect of this challenging partitioning scheme, we compare the IID-trained model to a model trained on the same dataset with natural partitioning. Qualitative results of the IID fine-tuned model are shown in Fig.~\ref{fig:pfhmr-visualization}.

\section{Local Data Annotation}
\label{sec:local_annotation}

Local data annotation, as described in Sec. 3.1 of the main paper, plays a critical role in improving the performance of \model with client-side user data.
Although there are many existing annotation approaches~\citep{bogo2016keep, pavlakos2019expressive, zioulis2023kbody, lin2024motion}, we choose a popular method, SMPLify-X~\citep{pavlakos2019expressive}, as the backbone of our local data annotation. SMPLify-X is an optimization-based human mesh regressor, where a high-quality SMPL \citep{loper2015smpl} mesh of $6890$ vertices can be obtained as the \textit{pseudo-ground-truth mesh}. The input 2D keypoints and depth supervision information for DePoser optimization are obtained through Sapiens~\citep{khirodkar2024sapiens}.

The model is pre-trained using the \model framework for 2000 global rounds with 100 clients and 10 clients per round under an IID partition. We then fine-tune the model with the locally annotated data using a learning rate of $1 \times 10^{-4}$ for 100 epochs over the user's local training split, with other training settings such as optimizers consistent with the centralized baseline.

\paragraph{Remark.} Notably, we currently employ an optimization-based approach to obtain pseudo-ground truth from monocular images. However, end devices possessing multi-view cameras could benefit from the additional 3D spatial information extracted by multiple cameras or depth sensors, thereby enhancing the local data annotation process using actual depth information with higher accuracy.

\section{Implementation Details}
\label{sec:imp_details}
\begin{figure}
\centering
\begin{overpic}[width=0.6\linewidth]{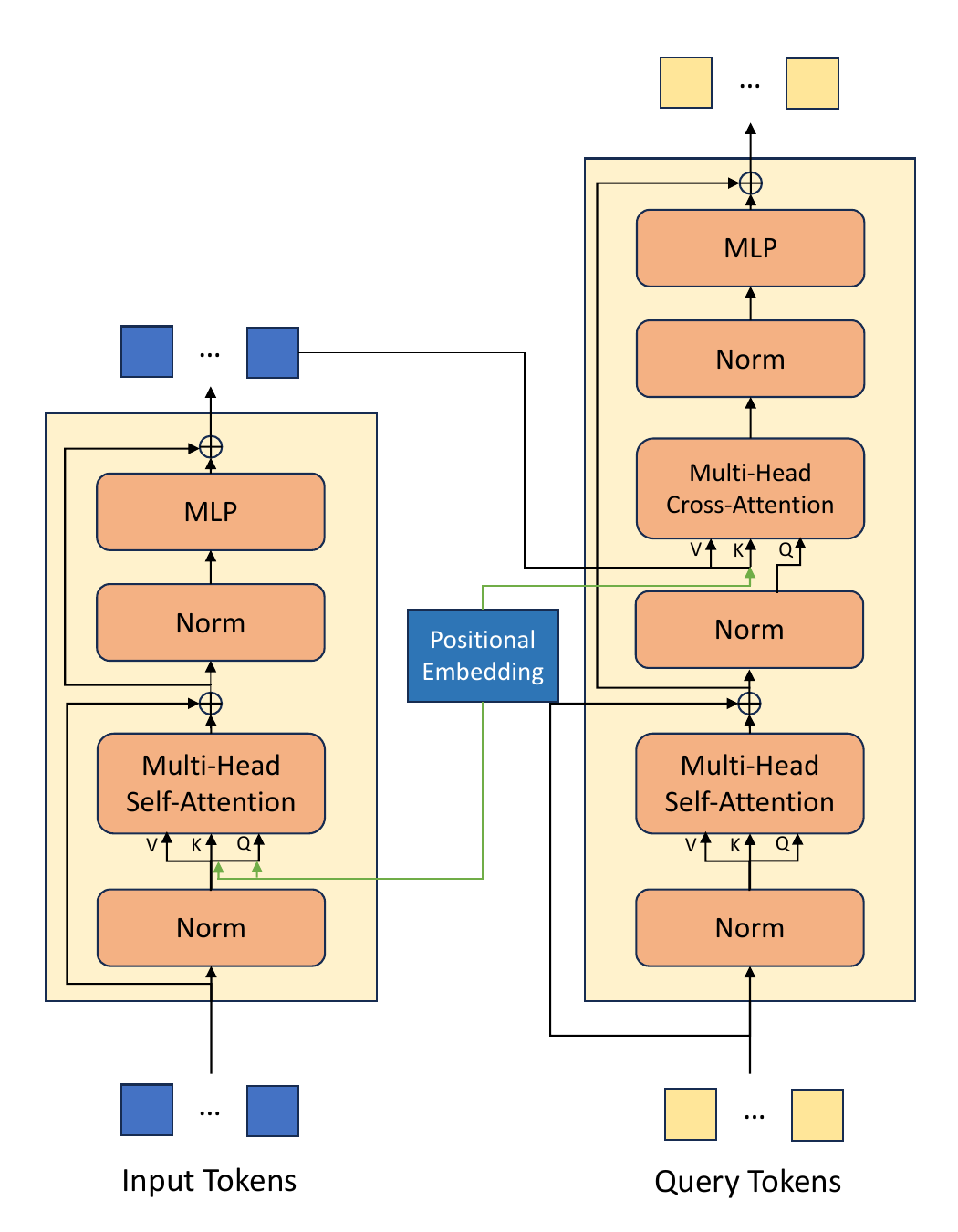}
\end{overpic}
\captionsetup{font=small,labelfont=bf}
\caption{Details of the Transformer encoder-decoder architecture used in FastMETRO \citep{cho2022cross} and TORE \citep{dou2023tore}.}
\label{fig:enc_dec_architecture}
\vspace{-5mm}
\end{figure}

In our experiments, we develop and evaluate our model based on two backbones including FastMETRO~\citep{cho2022cross} and TORE~\citep{dou2023tore} using the Transformer encoder-decoder architecture.
The details of this architecture are shown in Fig. \ref{fig:enc_dec_architecture}. Please refer to the original papers for further details.

\section{Discussion on MPC and DP}
\label{app:mpc_dp}
\begin{table}[t]
\centering
\small
\captionsetup{labelfont=bf}
\begin{tabular}{lccc}
\toprule
\textbf{Methods} & \textbf{Model} & \textbf{Round Time} & \textbf{Convergence} \\
\midrule
FL & TORE & 139.8s & Converge \\
MPC & TORE & 265.5s & Converge \\
DP & TORE & 139.8s & Non-Converge \\
\bottomrule
\end{tabular}
\vspace{-1mm}
\captionsetup{font=small,labelfont=bf}
\caption{Comparison of different privacy-preserving machine learning paradigms. The experiment uses the \texttt{10/10} TORE setting on an IID Human3.6M partition and reports average per-round runtime.}

\label{tab:mpc_dp}
\end{table}

We present a simple ablation setting comparing MPC and DP for the HMR model. For simplicity, we select the simplest \texttt{10/10} setting, i.e. data partitioned across 10 clients with all 10 sampled each round, with the TORE model using an IID-partitioned Human3.6M training set. For MPC, we choose to use secure aggregation, which implements MPC at the aggregation stage and hides individual client updates from an honest-but-curious server before aggregation. This acts as a baseline setting to demonstrate the computation and communication overhead of the MPC protocol.

For DP, we choose to account with R\'enyi Differential Privacy~\citep{mironov2017renyi}. We implement this with fixed clipping noise injection techniques. With the size of the Human3.6M dataset, we set $\delta = 1\times 10^{-6}$ and calculate the respective $\epsilon$ privacy budget together with the noise multiplier set. Following the suggestion on setting the L2 norm~\citep{bu2023automatic}, we set the L2 norm to a small value, starting with $1 \times 10^{-6}$.

The results are presented in Tab.~\ref{tab:mpc_dp}. As demonstrated by the average round time, which is almost double that of FL, the MPC approach shows significant communication and computation overhead.
However, the evaluation metrics (MPJPE, PA-MPJPE) still show similar performance with MPC, suggesting that MPC may be feasible to implement in HMR real-world settings with future improvement in hardware and communication.
Even worse, although we attempted an extensive search over hyperparameters, we could not get our large HMR model to converge even with a small noise scale multiplier ($\eta = 1$), resulting in a large $\epsilon = 199.7$ that offers a relatively small example-level DP guarantee. Due to the size of SOTA models such as TORE, the curse of dimensionality likely makes convergence hard, as the injected DP noise grows with the number of trained parameters~\citep{liu2024differentially}, and how to pre-train a large-scale model with DP efficiently remains an open question for future research.

\section{Discussion on Single-Device Training}
\label{app:naive-single-device-training}

We investigate the performance of na\"ive on-device training compared with privacy-constrained collaborative training under \model. We simulate single-device training with a small portion of the training dataset, and compare the accuracy with models trained with full data under centralized and privacy-constrained frameworks.

\begin{table}[t]
\centering
\small
\resizebox{0.7\columnwidth}{!}{%
\begin{tabular}{lccc|c}
\toprule
\textbf{Settings} & \textbf{Model} & \textbf{MPJPE} & \textbf{PA-MPJPE} & \textbf{Raw Data Local?} \\
\midrule
Centralized Baseline & FM-S & 57.98 & 40.62 &\xmark\\ \midrule
Na\"ive 5\% Data Training & FM-S & 83.78 & 63.65 & - \\
Na\"ive 10\% Data Training & FM-S & 73.04 & 53.36 & - \\
Na\"ive 20\% Data Training & FM-S & 65.10 & 47.01 & - \\ \midrule
\model 100\% Data Training & FM-S & 59.31 & 41.48 & \cmark\\
\midrule
Centralized Baseline & TORE-S & 63.88 & 41.99 &\xmark\\ \midrule
Na\"ive 5\% Data Training & TORE-S & 101.59 & 82.82 & - \\
Na\"ive 10\% Data Training & TORE-S & 84.82 & 63.52 & - \\
Na\"ive 20\% Data Training & TORE-S & 77.87 & 52.37 & -\\ \midrule
\model 100\% Data Training & TORE-S & 62.92 & 43.11 &\cmark\\
\bottomrule
\end{tabular}
}

\captionsetup{font=small,labelfont=bf}
\caption{\textbf{Comparison with Na\"ive Single-Device Training.} This table presents the performance metrics (MPJPE, PA-MPJPE) for training using various sizes (5, 10, 20\%) of the Human3.6M dataset to assess the on-device training scenario using only local data. Compared with na\"ive local training, which suffers from poor performance due to data scarcity, our \model could mitigate both issues due to the inter-device parameter sharing, achieving much higher performance while keeping raw data local. The \model in this table is trained in a distributed setting with 100 devices and 10 devices selected at random each time.}
\label{tab:pre-train-fl-finetune-table}
\end{table}

In Tab. \ref{tab:pre-train-fl-finetune-table}, we pre-train the model using $5,10,20\%$ of the Human3.6M dataset, which results in a large performance gap for both FastMETRO and TORE implementations. With 5\% of the data, na\"ive training is 23--26 mm worse than centralized training in MPJPE and PA-MPJPE for FastMETRO, and 38--41 mm worse for TORE; even with 20\% of the data, the gap remains 6--14 mm. Compared to the na\"ive single-device training setting, experiment results have confirmed the superiority of a larger training corpus, i.e. the full Human3.6M dataset.

\section{Discussion on Multiple Local Epochs}

\label{app:local_epoch}

\begin{table}[htbp]
\centering
\small
\captionsetup{labelfont=bf}

\resizebox{0.7\columnwidth}{!}{%
\begin{tabular}{cccc|c}
\toprule
\textbf{Local Epochs ($E$)} & \textbf{Model} & \textbf{MPJPE} & \textbf{PA-MPJPE} & \textbf{Raw Data Local?} \\
\midrule
1 & FM-S & 59.31 & 41.48 &\cmark\\
2 & FM-S & 59.28 & 42.42 &\cmark \\
3 & FM-S & 60.49 & 42.43 &\cmark \\
5 & FM-S & 59.52 & 42.07 &\cmark \\
10 & FM-S & 60.03 & 42.96 &\cmark \\
\midrule
1 & TORE-S & 62.92 & 43.11 &\cmark\\
2 & TORE-S & 61.84 & 43.27 &\cmark \\
3 & TORE-S & 63.53 & 42.74 &\cmark \\
5 & TORE-S & 63.41 & 43.18 &\cmark \\
10 & TORE-S & 63.88 & 43.39 &\cmark \\
\bottomrule
\end{tabular}
}
\captionsetup{font=small,labelfont=bf}
\caption{\textbf{Effect of Multiple Local Epochs.} Ablation on training the HMR model with more than one local epoch $E$ per global round in the federated setting. Note that raising $E$ raises the number of local optimizer steps each sampled client takes per round, and hence the total training budget.}
\label{tab:local-epochs}
\end{table}

We also conduct an ablation on model performance with multiple local epochs, i.e. letting each sampled client make more than one pass over its own shard before its update is aggregated. The experiment is presented with the \texttt{100/10} IID setting for the H36M dataset, i.e. data partitioned across 100 clients with 10 sampled each round.

As shown in Tab.~\ref{tab:local-epochs}, we found that, with more local epochs, the local model tends to overfit and results in less generalization capability and worse performance.

\section{Discussion on Data Distribution}
\label{app:data_distribution}

Heterogeneous data distributions are common in real-world privacy-constrained model training since each user's local data distribution may be different due to user identity, environment, and user habits. We therefore evaluate \model under heterogeneous data distributions. This section discusses each distribution used in \cref{sec:skewed-data-distribution} in more detail.

\subsection{LDA Data Partition}
\label{app:lda-data-partition}

One of the most utilized methodologies in FL to generate non-IID data is Latent Dirichlet Allocation (LDA)~\citep{hsu2019measuring}. Specifically, LDA works by drawing the distribution $\boldsymbol{q}$ over $N$ category classes of labels from a Dirichlet distribution, i.e. $\boldsymbol{q} \sim \mathrm{Dir}(\alpha \boldsymbol{p})$ where $\alpha$ is the concentration parameter controlling the skewness of the distribution and $\boldsymbol{p}$ is the vector characterizing the prior distribution over $N$ classes. This creates skewed datasets with an approximation to real-world non-IID distribution.

However, in our case, the label category for the training set is hard to determine. In particular, each human mesh is, in fact, a unique label by definition, i.e. they all represent unique annotations of their character. Hence, two alternative methods for considering pseudo-labeling are labeling by subjects and scenes or labeling by the dataset composition. As many datasets only have one scene per object (in particular, pseudo-3D labeled datasets), labeling by scene and subjects seems infeasible. Therefore, we opted to label it with dataset composition instead and consider the category class as the sub-dataset from which the image originated.

\begin{figure}[t]
 \centering
 \begin{subfigure}[b]{\columnwidth}
 \centering
 \includegraphics[width=0.7\textwidth]{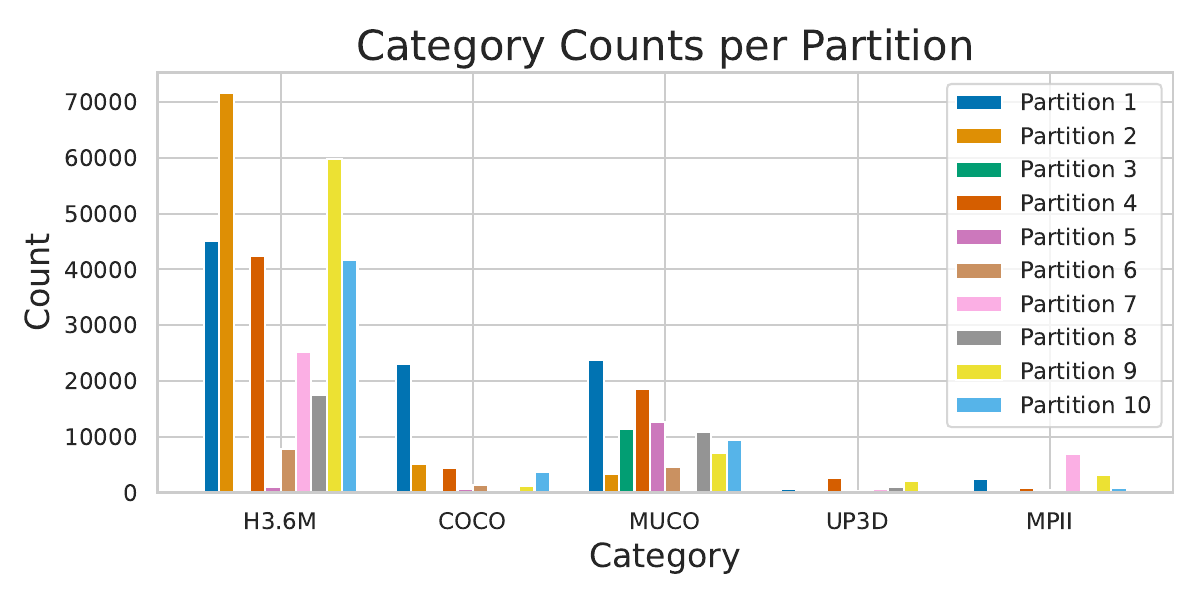}
 \caption{LDA Data Distribution for $\alpha=0.5$}
 \label{fig:lda-alpha-dist-05}
 \end{subfigure}
 \hfill
 \begin{subfigure}[b]{\columnwidth}
 \centering
 \includegraphics[width=0.7\textwidth]{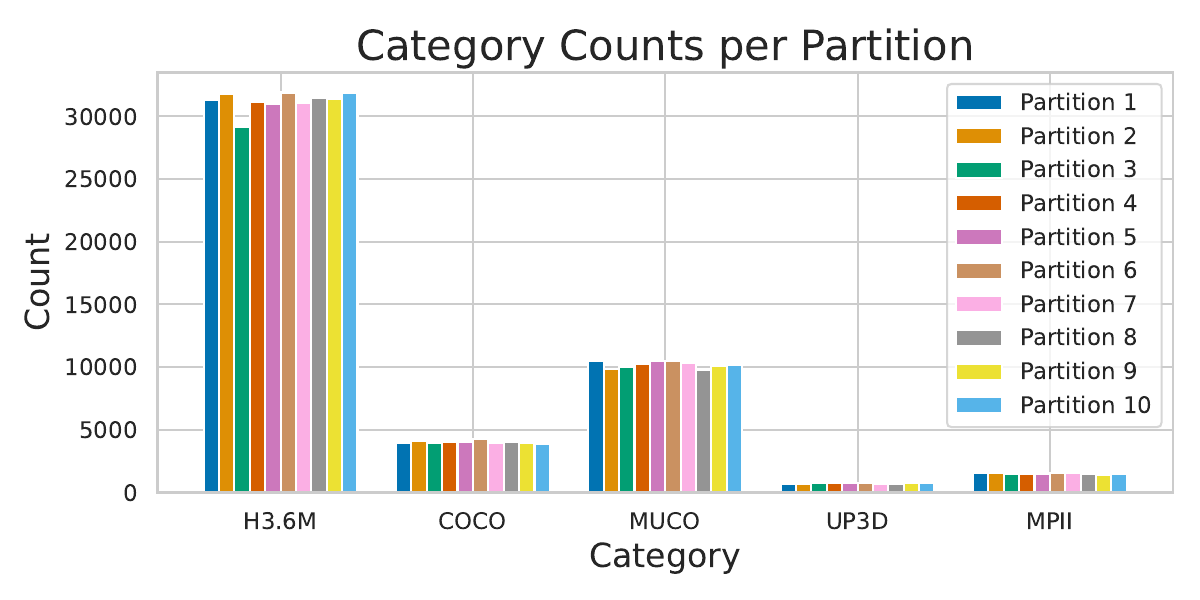}
 \caption{LDA Data Distribution for $\alpha=1000$}
 \label{fig:lda-alpha-dist-1000}
 \end{subfigure}
 \hfill
 \captionsetup{font=small,labelfont=bf}
 \caption{\textbf{Data Distribution of Applying LDA.} Distribution of labels from each dataset using LDA with 5 clients under (a) $\alpha=0.5$ and (b) $\alpha = 1000$. Note that the high $\alpha$ LDA retained a similar label distribution to the original category distribution, whereas the lower $\alpha$ introduced a much different category distribution.}
 \label{fig:lda-alpha-dist}
\end{figure}

On an IID dataset, LDA with high $\alpha$ yields a distribution similar to uniform sampling, whereas low $\alpha$ yields a much more skewed non-IID distribution. Our training dataset is already non-IID by composition, so LDA partitioning only increases its skewness: low $\alpha$ makes it much more skewed, whereas high $\alpha$ keeps it close to the original dataset composition (Fig.~\ref{fig:lda-alpha-dist}).

\begin{figure}[htbp]
 \centering
 \begin{subfigure}[b]{\columnwidth}
 \centering
 \includegraphics[width=0.7\textwidth]{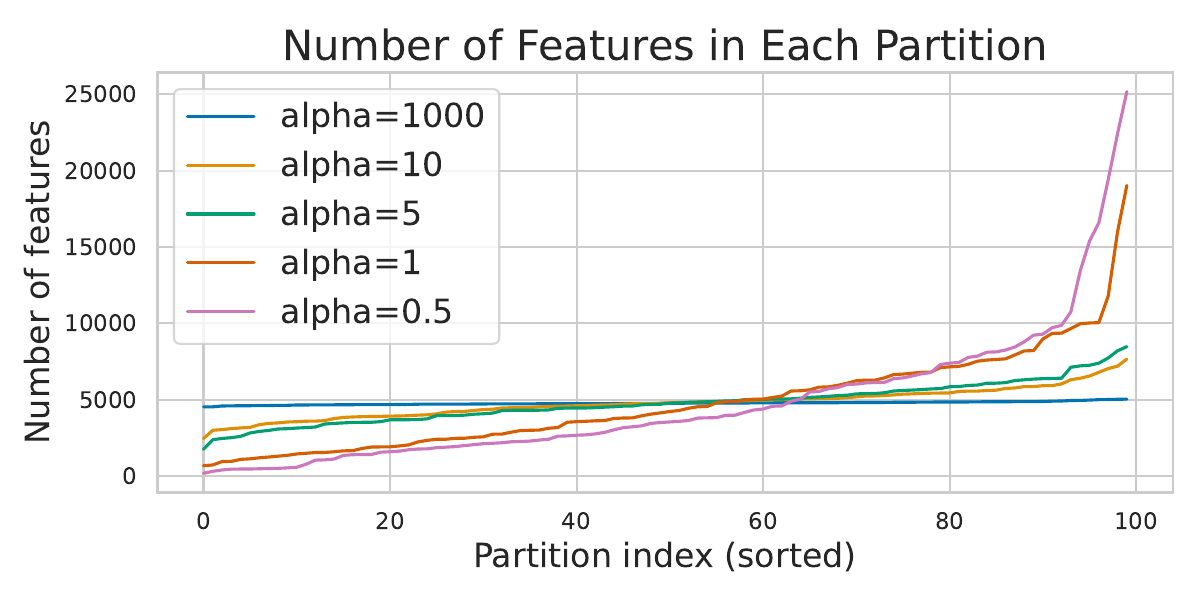}
 \caption{Comparison of Data Assignment in LDA with 100 Partitions}
 \label{fig:lda-partition-num-100}
 \end{subfigure}
 \hfill
 \begin{subfigure}[b]{\columnwidth}
 \centering
 \includegraphics[width=0.7\textwidth]{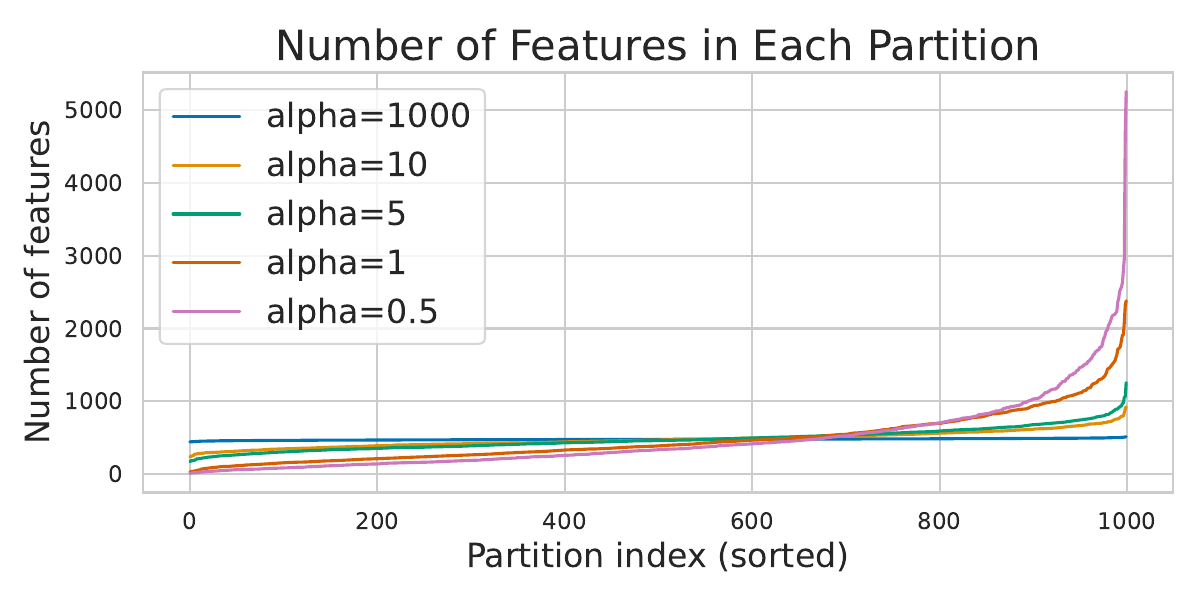}
 \caption{Comparison of Data Assignment in LDA with 1000 Partitions}
 \label{fig:lda-partition-num-1000}
 \end{subfigure}
 \hfill
 \captionsetup{font=small,labelfont=bf}
 \caption{\textbf{LDA-Partitioned Dataset Size for Individual Clients.} This figure illustrates the size of the dataset assigned to each client, sorted by the size of data for (a) 100 partitions and (b) 1000 partitions, using LDA. Note that the lower $\alpha$ creates a much more skewed long-tail distribution following the Dirichlet distribution that it is sampled from. }
 \label{fig:lda-partition-num}
\end{figure}

In addition, as shown in Fig.~\ref{fig:lda-partition-num}, LDA also creates the client dataset size distribution following the Dirichlet distribution, i.e. lower $\alpha$ creates a much more skewed long-tail distribution in the number of samples assigned to each client. This variation in client dataset size mirrors real deployments, where clients hold very different amounts of data.

\subsection{CNN Activation Clustering}
\label{app:cnn-cluster}
\begin{figure}[h]
 \centering
 \includegraphics[width=0.7\columnwidth]{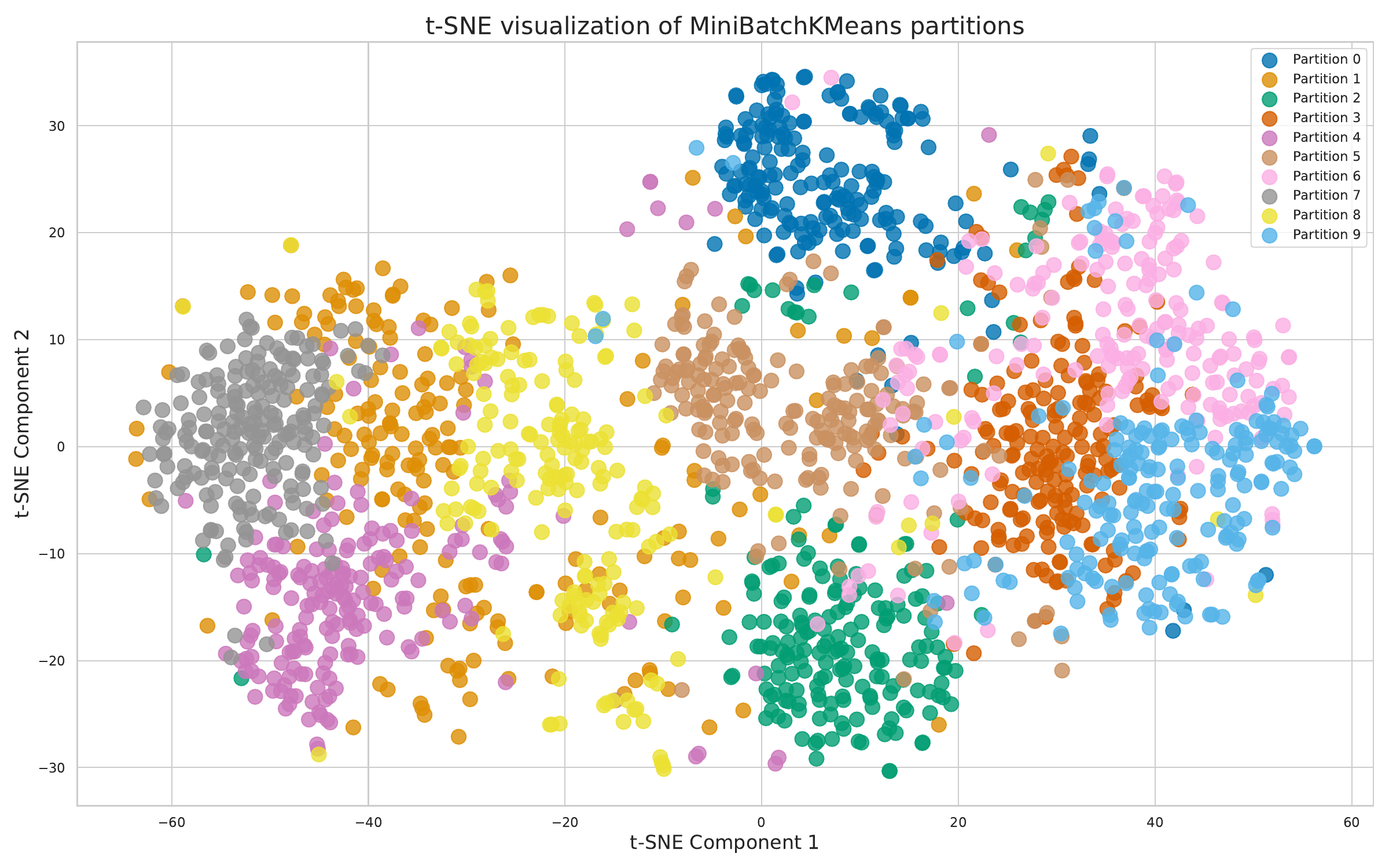}
 \captionsetup{font=small,labelfont=bf}
 \caption{\textbf{CNN-Partitioned Data Visualization.} For visualization, this diagram clusters the selected datasets into 10 partitions with \texttt{MiniBatchKMeans} (the experiments use 100) and plots them using t-SNE~\citep{van2008visualizing}. Each partition is sampled with 100 data points for ease of computation. The data points with the same color come from the same cluster. }
 \label{fig:cnn-partition-distribution}
\end{figure}

Another approach to pseudo-partitioning is considering natural clustering from the features. We use the same CNN feature extractor (ResNet-50; \citealp{he2016deep}) from the \model\ model to extract the feature from the last-layer activation. The feature $x_{out}$ extracted from the last-layer activation is then clustered into 100 pseudo-partitions using \texttt{MiniBatchKMeans}~\citep{sculley2010web}. An example of the effects of clustering is illustrated in Fig.~\ref{fig:cnn-partition-distribution}.

Upon inspection, we find that the feature extractor clusters also mimic the natural dataset composition relationship of the underlying dataset: the large groups of the clusters on the left consist mostly of the Human3.6M dataset, whereas the large cluster on the right consists of the pseudo-3D labeled dataset within the training dataset mixture. Clustering on CNN features therefore yields a partition that differs from IID sampling.

Under this setting, as shown in Tab. \ref{tab:data-dist-fl-table}, our TORE \citep{dou2023tore} model achieves the same level of accuracy as the centralized model, with a lower MPJPE error and slightly higher (+0.26 mm) PA-MPJPE error. The FastMETRO \citep{cho2022cross} variant also performs on par with the centralized model, achieving a lower MPJPE and slightly higher PA-MPJPE (+0.21 mm).

\subsection{Natural Partitioned In-the-wild Dataset}
\label{app:3dpw-natural-partition}

\begin{figure}[htbp]
 \centering
 \includegraphics[width=0.9\columnwidth]{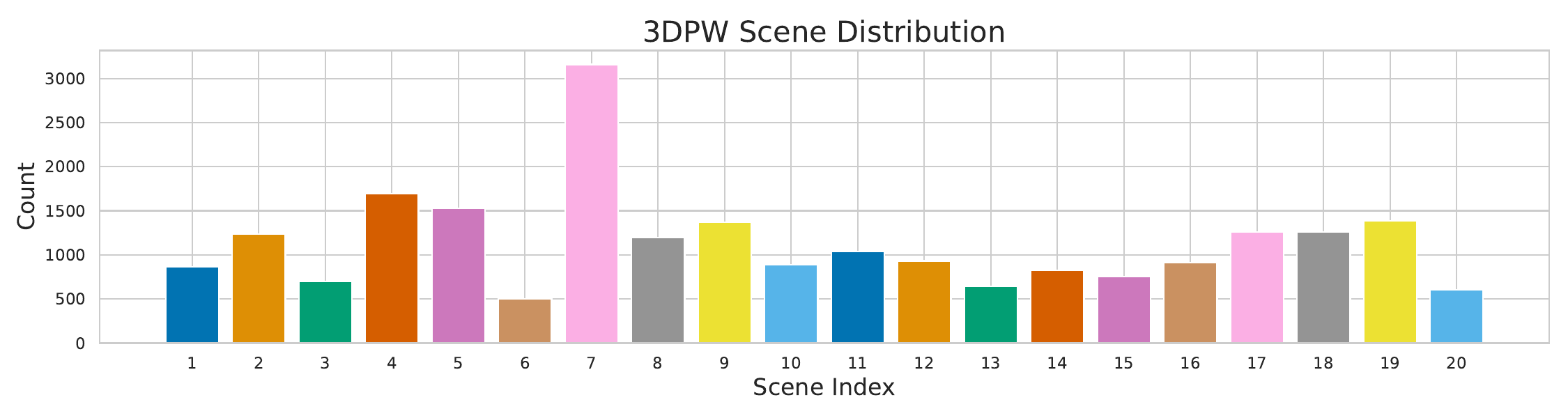}
 \captionsetup{font=small,labelfont=bf}
 \caption{\textbf{3DPW Dataset Distribution.} This diagram shows the natural distribution of different scenes in the 3DPW~\citep{von2018recovering} dataset. Note that this is only the train set of the 3DPW dataset following the established data train-val partition. }
 \label{fig:3dpw-dataset-distribution}
\end{figure}

In addition, the centralized HMR model is usually trained on synthetic and indoor datasets due to the ease of gathering, not on the sparsely collected data from in-the-wild applications. These data usually exhibit a much different data distribution. We aim to further evaluate \model on this real-world dataset with naturally partitioned heterogeneous distribution labels. We utilize 3DPW~\citep{von2018recovering} as an example of such real-world collected in-the-wild data that presents much more challenging settings and is naturally partitioned with scene labels, which is demonstrated in Fig.~\ref{fig:3dpw-dataset-distribution}. This makes them a good representation of the natural heterogeneous distribution that \model deployment is likely to face in real-world applications.

\subsection{Heterogeneous Distribution Results}

\begin{table}[h!]
\centering
\small
\resizebox{0.7\columnwidth}{!}{%
\begin{tabular}{lccc|c}
\toprule
\textbf{Settings} & \textbf{Model} & \textbf{MPJPE} & \textbf{PA-MPJPE} &\textbf{Raw Data Local?} \\
\midrule
Centralized & FM-S & 57.98 & 40.62 &\xmark\\ \midrule
$\alpha$ = 0.1 & FM-S & 51.11 & 37.24 & \cmark\\
$\alpha$ = 0.5 & FM-S & 57.80 & 39.82 &\cmark\\
$\alpha$ = 1 & FM-S & 59.63 & 39.80 &\cmark\\
$\alpha$ = 10 & FM-S & 56.09 & 39.84 &\cmark\\
$\alpha$ = 100 & FM-S & 57.53 & 40.69 &\cmark\\
$\alpha$ = 1000 & FM-S & 59.80 & 41.80 &\cmark\\
CNN Cluster & FM-S & 56.84 & 40.83 & \cmark\\
\midrule
Centralized & TORE-S & 63.88 & 41.99 &\xmark\\ \midrule
$\alpha$ = 0.1 & TORE-S & 57.02 & 39.10 & \cmark\\
$\alpha$ = 0.5 & TORE-S & 58.91 & 40.94 & \cmark\\
$\alpha$ = 1 & TORE-S & 62.39 & 41.76 & \cmark\\
$\alpha$ = 10 & TORE-S & 60.96 & 42.25 &\cmark\\
$\alpha$ = 100 & TORE-S & 63.08 & 42.59 &\cmark\\
$\alpha$ = 1000 & TORE-S & 63.74 & 42.50 &\cmark\\
CNN Cluster & TORE-S & 63.14 & 42.25 &\cmark\\

\bottomrule
\end{tabular}
}
\captionsetup{font=small,labelfont=bf}
\caption{\textbf{Data Distribution Evaluation.} MPJPE and PA-MPJPE under different data distributions (LDA with varying $\alpha$ and CNN clustering), evaluated on Human3.6M~\citep{ionescu2013human3}. \model's error is competitive with the centralized baseline in every setting and lower in some skewed partitions; we do not isolate the cause of these gains. All models (except centralized) in this table are trained in a distributed setting with 100 devices, with 10 devices selected at random each round. }
\label{tab:data-dist-fl-table}
\end{table}

Tab.~\ref{tab:data-dist-fl-table} lists the full results of the heterogeneous experiments.

\section{Performance of 3DPW Fine-tuning}
\label{app:3dpw-finetune}

\begin{figure}
\centering
\begin{overpic}[width=0.55\linewidth]{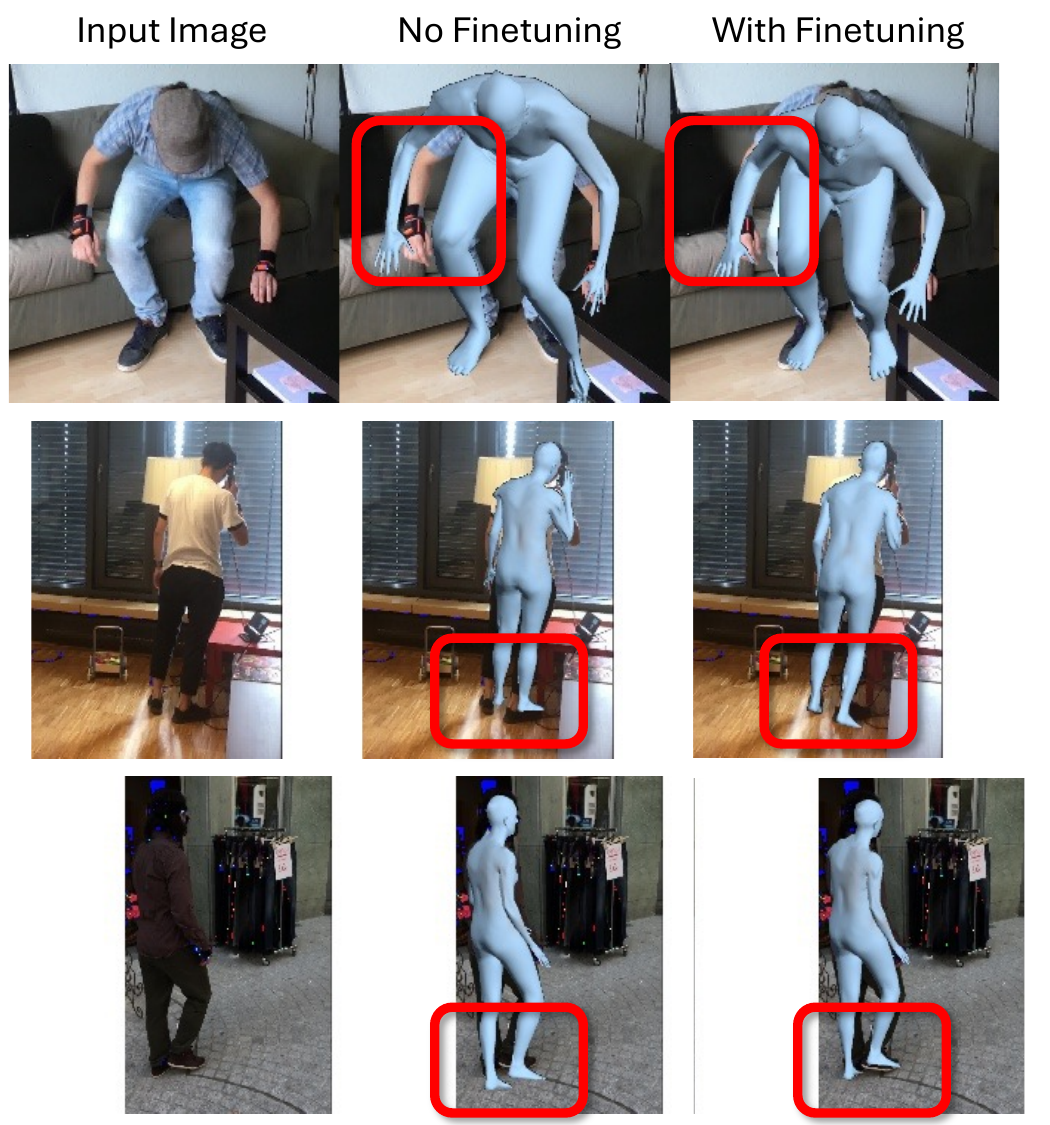}
\end{overpic}
\captionsetup{font=small,labelfont=bf}
\caption{Qualitative results of \citet{cho2022cross} before and after privacy-constrained fine-tuning on the 3DPW \citep{von2018recovering} dataset using the \model framework.}
 \label{fig:pf_finetuning}
\vspace{-5mm}
\end{figure}

One potential use case of the \model framework is performing privacy-constrained on-device fine-tuning to align the model with real-world data. We demonstrate this by pre-training our model on a series of datasets \citep{ionescu2013human3, lassner2017unite, lin2014microsoft, andriluka20142d, mehta2018single} and using \model to fine-tune on 3DPW \citep{von2018recovering} with 100 clients and 10 clients randomly sampled per round. As shown in Tab. \ref{tab:privacy-friendly-finetuning}, both fine-tuned models reduce MPJPE by about 32 mm compared to the model without fine-tuning. Fig. \ref{fig:pf_finetuning} also gives qualitative evidence for the on-device fine-tuning, where the reconstructed mesh accuracy visibly improves. Privacy-constrained fine-tuning stays within 1.3 mm of centralized fine-tuning on every metric while keeping raw data on the device.

\begin{table}[htb]
\centering
\small
\resizebox{\columnwidth}{!}{%
\begin{tabular}{lcccc|c}
\toprule
\textbf{Setting} & \textbf{Model} & \textbf{MPJPE} & \textbf{PA-MPJPE} & \textbf{MPVPE} & \textbf{Raw Data Local?} \\
\midrule
No Fine-tuning & FM-S & 116.35 & 70.20 & 132.96 &\xmark\\
Centralized Fine-tuning & FM-S & 84.92 & 54.69 & 97.6 &\xmark\\ \midrule
\model Fine-tuning & FM-S & 84.66 & 54.78 & 97.42 &\cmark\\
\midrule
No Fine-tuning & TORE-S & 119.64 & 68.35 & 138.93 &\xmark\\
Centralized Fine-tuning & TORE-S & 87.97 & 55.35 & 101.88 &\xmark\\ \midrule
\model Fine-tuning & TORE-S & 87.55 & 54.08 & 101.86 & \cmark\\
\bottomrule
\end{tabular}
}
\captionsetup{font=small,labelfont=bf}
\caption{\textbf{Privacy-Constrained Fine-Tuning Table.} Results of privacy-constrained fine-tuning of models \citep{dou2023tore, cho2022cross} pre-trained on Human3.6M, UP-3D, COCO, MPII and MuCo-3DHP~\citep{ionescu2013human3, lassner2017unite, lin2014microsoft, andriluka20142d, mehta2018single} using \model on the 3DPW dataset \citep{von2018recovering}, compared with centralized fine-tuning with the same datasets and the model without fine-tuning. The privacy-constrained fine-tuning is conducted with 100 clients and 10 randomly sampled per round.}
\label{tab:privacy-friendly-finetuning}
\end{table}

\section{On-Device Performance}
\label{app:on-device-perf}

We further assess the feasibility of the pipeline by measuring the performance of our \model on a real edge device. We selected one of the representative embedded devices, Nvidia AGX Orin~\citep{nvidia2021nvidia}, to run 5 rounds as one client in the 100- and 1000-client settings, following the experiment setting as outlined in the previous sections. This single-device measurement indicates the runtime scale of edge deployment; a full multi-client deployment may behave differently. With more capable hardware being introduced in applications such as VR and AR, we expect the runtime profile to improve on newer hardware in the coming years.

\begin{table}[htbp]
\vspace{-3mm}
 \centering
 \resizebox{0.8\columnwidth}{!}{%
 \begin{tabular}{cccccc}
\toprule
\multirow{2}{*}{\textbf{Model}} & \multirow{2}{*}{\textbf{Device}}
& \multicolumn{2}{c}{\textbf{100 Partitions}} & \multicolumn{2}{c}{\textbf{1000 Partitions}} \\ \cmidrule(lr){3-6}
 & & \textbf{Throughput} & \textbf{Time} & \textbf{Throughput} & \textbf{Time} \\
\midrule
FM-S & Jetson AGX - 15W & 2.46 & 1629.33 & 2.42 & 165.33 \\
FM-S & Jetson AGX - 50W & 8.01 & 499.67 & 7.81 & 51.23 \\
FM-S & RTX A6000 Ada & 56.59 & 70.7 & 48.0 & 8.33 \\
\midrule
TORE-S & Jetson AGX - 15W & 3.43 & 1166.06 & 3.35 & 119.29 \\
TORE-S & Jetson AGX - 50W & 8.65 & 461.5 & 8.42 & 47.49 \\
TORE-S & RTX A6000 Ada & 50.87 & 78.67 & 44.40 & 9.02 \\
\bottomrule
\end{tabular}
}
\captionsetup{font=small,labelfont=bf}
 \caption[On-Device Performance Table]{\textbf{On-Device Performance Table.} This table shows the image throughput and total time used for training with 100 (4000 images) and 1000 (400 images) partitions of client data on different power profiles for Nvidia AGX Orin. We also compare the image throughput and total time with a desktop-level graphics card (RTX A6000 Ada). The unit for Throughput is images per second (img/s), and the unit for Time is seconds (s).}
 \label{tab:on-device-perf-table}
\vspace{-2mm}
\end{table}

The runtime and throughput results are shown in Tab.~\ref{tab:on-device-perf-table}. In the 1000-partition configuration, the edge device would need to train for 85 hours under the 50W power configuration. Nonetheless, the communication cost compared with both figures is relatively insignificant (42 seconds per round), assuming a 37.4 Mb/s broadband upload speed, which follows the median uplink speed in the US \citep{speedtest2024united} between June and August 2024. \model's runtime is therefore within the range of practical cross-device FL systems, whose training can take hundreds of hours~\citep{huba2022papaya, bonawitz2019towards}.

\end{document}